\documentclass{article}
\usepackage{arxiv}
\usepackage[linesnumbered,ruled,vlined]{algorithm2e}
\usepackage{amsmath}
\usepackage{enumitem}
\usepackage[utf8]{inputenc}
\usepackage[T1]{fontenc}
\usepackage{hyperref}
\usepackage{url}
\usepackage{booktabs}
\usepackage{amsfonts}
\usepackage{nicefrac}
\usepackage{microtype}
\usepackage{cleveref}
\usepackage{lipsum}
\usepackage{graphicx}
\usepackage{makecell}
\usepackage[square,numbers,sort&compress]{natbib}
\usepackage{doi}
\usepackage{float}
\usepackage{multirow}
\usepackage{rotating}
\usepackage{tabularx}
\usepackage{booktabs}
\usepackage{amssymb}
\newcolumntype{Y}{>{\raggedright\arraybackslash}X}
\newcolumntype{Z}{>{\centering\arraybackslash}X}
\usepackage{appendix}
\usepackage{soul, xcolor}

\title{Collaborative Optimization of Multiclass Imbalanced Learning: Density-Aware and Region-Guided Boosting}

\newif\ifuniqueAffiliation
\uniqueAffiliationfalse

\ifuniqueAffiliation

\else
\usepackage{authblk}

\author[1,2]{Chuantao Li\textsuperscript{\dag}}
\author[1]{Zhi Li\textsuperscript{\dag}}
\author[1]{Jiahao Xu}
\author[1]{Jie Li}
\author[1]{Sheng Li\thanks{
    Corresponding author.\\
    \textsuperscript{\dag}These authors contributed equally to this work.\\
    \textit{E-mail addresses:} 11672411dd@stu.gdou.edu.cn (C. Li), lizhi@gdou.edu.cn (Z. Li), xujh284@stu.gdou.edu.cn (J. Xu), lj@stu.gdou.edu.cn (J. Li), lish\_ls@gdou.edu.cn (S. Li)}}

\affil[1]{School of Automation Engineering, University of Electronic Science and Technology of China, Chengdu 611731, Sichuan, China}
\affil[2]{School of Mathematics and Computer, Guangdong Ocean University, Zhanjiang 524088, Guangdong, China}

\fi
\renewcommand{\shorttitle}{Density-Aware and Region-Guided Boosting}

\hypersetup{colorlinks,linkcolor=blue,citecolor=blue,urlcolor=blue}
\hypersetup{
pdftitle={Collaborative Optimization of Multiclass Imbalanced Learning: Density-Aware and Region-Guided Boosting},
pdfsubject={cs.LG},
pdfauthor={Chuantao Li, Zhi Li, Jiahao Xu, Jie Li, Sheng Li},
pdfkeywords={Imbalanced Learning, Collaborative Optimization, Ensemble Learning, Sample Weight Update},
}

\begin{document}
\maketitle

\begin{abstract}
Numerous studies on Boosting attempt to mitigate classification bias caused by class imbalance. However, existing studies have yet to explore the collaborative optimization of imbalanced learning and model training. This constraint hinders further performance improvements. To bridge this gap, this study proposes a collaborative optimization Boosting model of multiclass imbalanced learning. By integrating the density factor and the confidence factor, this model implements a noise-resistant weight update mechanism alongside a dynamic sampling strategy. Rather than functioning as independent components, these modules are tightly integrated to orchestrate weight updates, sample region partitioning, and region-guided sampling. Thus, this study proposes the collaborative optimization of imbalanced learning and model training. Extensive experiments on 40 public imbalanced datasets demonstrate that the proposed model significantly outperforms seven state-of-the-art baselines. The code and datasets for this paper are available at: \href{https://github.com/ChuantaoLi/DARG}{https://github.com/ChuantaoLi/DARG}.
\end{abstract}


\section{Introduction}

In industrial applications such as fault classification \cite{1}, medical diagnosis \cite{2}, network intrusion detection \cite{3}, and image recognition \cite{4}, data often exhibits a pronounced long-tail distribution. Such class imbalance poses a fundamental challenge in machine learning. Minority class samples contribute negligibly to the aggregate training loss, resulting in models that favor the majority class. Consequently, the classifier often fails to capture discriminative patterns in minority samples. While a model might achieve high overall accuracy, this metric is often misleading and of limited practical use, as the majority class dominates the predictions.

Researchers have developed numerous imbalanced learning methodologies over the past two decades, generally categorized into data-level and algorithm-level \cite{5, 6}. Data-level methodologies use resampling to rebalance the sample distribution, enhancing the model's ability to learn from minority classes. However, by sampling or removing samples, these methods risk causing overfitting, information loss, or class overlap \cite{7, 8, 9}. To improve the sampling quality, region-based strategies have become the prevalent paradigm. Unlike traditional methods like SMOTE \cite{7} and ADASYN \cite{10} that synthesize samples directly, recent studies prioritize initial data clustering followed by intra-cluster sampling. This ensures that generated samples more faithfully obey the original distribution \cite{11, 12, 13}.

In contrast, algorithm-level methodologies modify loss functions or misclassification costs to prioritize minority samples \cite{14, 15}. However, effectively configuring cost matrices or sample weights remains difficult. Furthermore, these sample weights typically remain static during training, making it difficult to optimize them collaboratively with the model. To address this, some researchers embed sampling modules into the Boosting process. By modifying the training distribution before each iteration, the base learner can treat the minority class more equitably, while the classification error rate of each base learner is calculated to update the sample weights \cite{16, 17, 18}.

Despite these advancements, inherent limitations still persist. In many datasets, no distinct decision boundaries exist between majority and minority classes. For data-level methodologies, isolated noise samples often degrade the quality of region partitioning and sample generation, blurring decision boundaries and inducing class overlap. For algorithm-level methodologies, most existing studies merely splice the sampling module into the Boosting process rather than achieving deep coupling. Additionally, most methods target binary classification and do not extend to multiclass scenarios.

We observe that the AdaBoost weight update mechanism provides a bridge to collaboratively optimize the sampling module and the Boosting process. Crucially, this synergy allows us to integrate the complementary perspectives of both methodologies: whereas data-level methodologies use neighbors or distances to quantify sample density and region information, algorithm-level methodologies use prediction probabilities and training states to evaluate sample difficulty. By incorporating density and confidence information into the weight update mechanism, it can guide the base learner to focus on high-value samples while mitigating noise interference. This approach remedies a key deficiency in traditional AdaBoost, which focuses indiscriminately on all misclassified samples. Furthermore, sampling module does not only provide a balanced distribution but also executes dynamic sampling tailored to each base learner and sample region based on the current training state.

Moving beyond a single research path, we focus on the cooperative optimization of imbalanced learning and model training. This study proposes Density-Aware and Region-Guided Boosting for multiclass imbalance. Our main contributions are:

\begin{enumerate}[label=(\arabic*), leftmargin=*]

\item \textbf{A noise-resistant sample weight update mechanism based on dual factors.} By devising a density factor based on mutual nearest neighbors and a confidence factor based on classification hardness,  the proposed mechanism distinguishes between noise and genuinely hard samples, assigning higher weights to high-value samples.
    
\item \textbf{A cooperative optimization model for sample region partitioning and weight updates.} Based on density factor and confidence factor, this study partitions minority class samples into dense, boundary, and noise regions. Moreover, this study integrates this with a progressive sampling scheduler to execute dynamic sampling. 

\item \textbf{A region-guided generation strategy for dynamic sampling.} By originating one end of the synthetic sample in a boundary region and the other in a dense region, the strategy pulls ambiguous samples toward high-confidence dense regions. This effectively mitigates class overlap and reconstructs distinct decision boundaries.

\end{enumerate}

The remainder of this paper is organized as follows. Section \ref{sec:Related Work} reviews the progress and limitations of related work in imbalanced learning. Section \ref{sec:Methodology} introduces the implementation of our proposed model. Section \ref{sec:Experiments and Discussion} reports the experiments and analyzes the results. Section \ref{sec:Conclusion} summarizes the proposed model and discusses future work.

\section{Related Work}
\label{sec:Related Work}

\subsection{Data-Level Methodologies}

Data-level methodologies mitigate class imbalance by either sampling minority class samples or undersampling majority class samples. SMOTE \cite{7}, the most foundational approach, generates new samples through stochastic linear interpolation between minority class samples. Augmenting the minority class improves model fitting. Nevertheless, random interpolation often results in low-quality synthetic samples. Several variants have since emerged to improve this process. Han et al. introduced Borderline-SMOTE, which restricts oversampling to minority samples near decision boundaries \cite{19}. He et al. proposed ADASYN, which adaptively allocates the oversampling rate based on the density of surrounding majority class samples \cite{10}. Barua et al. developed MWMOTE, utilizing a two-stage nearest neighbor search to weight informative boundary samples for oversampling \cite{20}. Sun et al. proposed EB-SMOTE, which expands the minority and majority borderlines toward each other to alleviate class overlap \cite{10902063}. While these methods improve upon basic interpolation, they often rely on K-nearest neighbors. When parent samples reside in regions dominated by the opposing class, these noise points receive excessive attention, leading to the generation of noisy synthetic samples. To suppress such noise, Li et al. proposed WRND, a weighted oversampling framework with relative neighborhood density for noisy classification \cite{LI2024122593}.

To address noise interference and class overlap, recent studies employ clustering and region partitioning strategies. Thakur et al. proposed CBReT, which uses a Gaussian Mixture Model to cluster minority samples and oversamples within each cluster \cite{11}. Furthermore, by calculating the distance between new samples and cluster centroids, this method retains only those samples within a reasonable range. Xie et al. introduced PUMD, which extracts majority class samples by computing density peaks, mitigating the information loss common in traditional cluster-based undersampling methods \cite{12}. Compared to distance-based strategies, clustering-based more accurately identifies the data distribution; noise samples are typically flagged as outliers or relegated to trivial clusters, preventing them from degrading the oversampling quality.

While effective, CBReT and PUMD primarily target binary classification. Multiclass tasks present more complex decision boundaries that these strategies often struggle to navigate. Consequently, Liu et al. proposed NROMM, which incorporates an adaptive safe embedding mechanism into intra-cluster generation \cite{21}. They further decompose clusters into safe, boundary, and trapped samples based on local density to apply region-specific generation. Shen et al. proposed MCNRO, utilizing a Gaussian Radial Basis Function to calculate class potential energy \cite{22}. By comparing potential energy levels across classes, they partition oversampling regions into high and low potential regions, better accommodating multiclass overlap. Additionally, Xia et al. introduced GBSMOTE, which groups samples using slack variables from the SVM training process \cite{23}. Their study argues that oversampling in a high-dimensional kernel space, rather than in the original geometric space, better alleviates the challenges posed by complex distributions.

Despite these enhancements, data-level methodologies remain limited by their reliance on the dataset's prior distribution. Specifically, these methods configure cluster distributions and partition thresholds based on the original static data. However, the designation of a sample as dense, boundary, or noise should ideally evolve alongside the training process. This lack of dynamic adjustment represents a significant bottleneck for data-level performance.

\subsection{Algorithm-Level Methodologies}

Algorithm-level methodologies encompass cost-sensitive learning and ensemble learning methods. Cost-sensitive methods handle the class imbalance problem by assigning different misclassification costs or sample weights to different classes. Wang et al. proposed AdaBoostNC, which introduces a differentiated cost matrix to penalize minority class misclassifications more heavily \cite{6170916}. Tang et al. developed TSBQT, a two-stage method that allocates adaptive costs based on classification difficulty to suppress label noise \cite{25}. However, designing reasonable cost matrices or sample weights remains a challenge, especially in multiclass problems \cite{18}. Therefore, compared with cost-sensitive methods, ensemble learning methods have received more favor in algorithm-level research.

Ensemble learning methods construct a strong learner by combining multiple base learners, offering robust generalization and flexibility. Early examples include SMOTEBoost \cite{make8070211} and RUSBoost \cite{26}, which balance class distributions in each iteration to attenuate the influence of the majority class. In recent years, these methods have been further developed. For example, Rodríguez et al. proposed MultiRandBal. Their study extends the Random Balance strategy to multiclass problems by stochastically assigning class ratios for each base classifier \cite{16}. Zhu et al. proposed OREMBoost. Their study embeds a clean sub-region discovery mechanism into the Boosting process to generate samples exclusively in safe regions \cite{17}. Li et al. proposed GRDSAD, a graph discriminative dynamic sampling AdaBoost for imbalanced cardiovascular disease diagnosis \cite{LI2026129546}. Although these improvement strategies have achieved better performance in multiclass imbalance problems, they do not fully exploit the characteristics of the Boosting process itself. These studies merely incorporate an improved sampling method into the base learners.

In contrast, some other research on ensemble learning have shifted toward adaptively adjusting the model's focus by enhancing the Boosting weight update mechanism. For example, Li et al. proposed AdaBoostAD. Their study integrates the between class imbalance ratio, the within class density variable, and the adaptive margin altogether into the AdaBoost framework. By dynamically adjusting sample weights, it augments the model's classification capability on multiclass imbalanced data \cite{18}. Datta et al. proposed LexiBoost, modeling the weight selection problem as a lexicographic linear programming game to balance majority and minority class tradeoffs \cite{8621023}. Jan et al. proposed JanEnsemble. Their study uses marginal distributions to estimate class preferences and dynamically adjust attention without manual weight configuration \cite{27}. These methods adjust the attention to the majority and minority classes, achieving an adaptive balance mechanism without manually configuring weights. By utilizing statistical information such as density, imbalance ratio, and confidence to furnish real-time feedback to the base learner, these methods are demonstrated substantial adaptability and dynamics.

\subsection{Limitations of Existing Methodologies}

In recent years, data-level methodologies have concentrated on how to enhance sampling quality. Although data-level methodologies have improved sampling quality by transitioning from nearest-neighbor to clustering strategies, these methods remain constrained by the static prior knowledge of the original data. Consequently, they cannot dynamically adjust based on the current state of the training model.

In contrast, algorithm-level methodologies embed imbalanced learning directly into the construction of strong learner. However, many studies merely balance the data before each training epoch. This rudimentary operation fails to synchronize the sampling process with the model’s internal optimization. Furthermore, although enhancing weight update mechanisms is beneficial, simply adjusting weights is often insufficient for complex multiclass datasets that lack distinct decision boundaries. Noise samples and class overlap continue to induce misclassifications that weight adjustment alone cannot surmount.

\section{Methodology}
\label{sec:Methodology}

\subsection{Model Overview}

To resolve the aforementioned challenges, this study proposes the cooperative optimization of imbalanced learning within the Boosting process. We improve the sample weight update mechanism by introducing a density factor and a confidence factor, mitigating the inherent limitation of traditional AdaBoost, which focuses exclusively on misclassified samples while ignoring their underlying data distribution or noise characteristics. These factors enable precise partitioning of samples into regions and the design of a progressive sampling scheduler. This scheduler dynamically adjusts the number of samples in each epoch, thereby aligning imbalanced learning with model training. To mitigate noise interference and class overlap, we employ a region-guided generation strategy that pulls ambiguous boundary samples toward high-confidence dense regions, effectively reconstructing distinct decision boundaries. 

\begin{figure}
    \centering
    \includegraphics[width=1.0\linewidth]{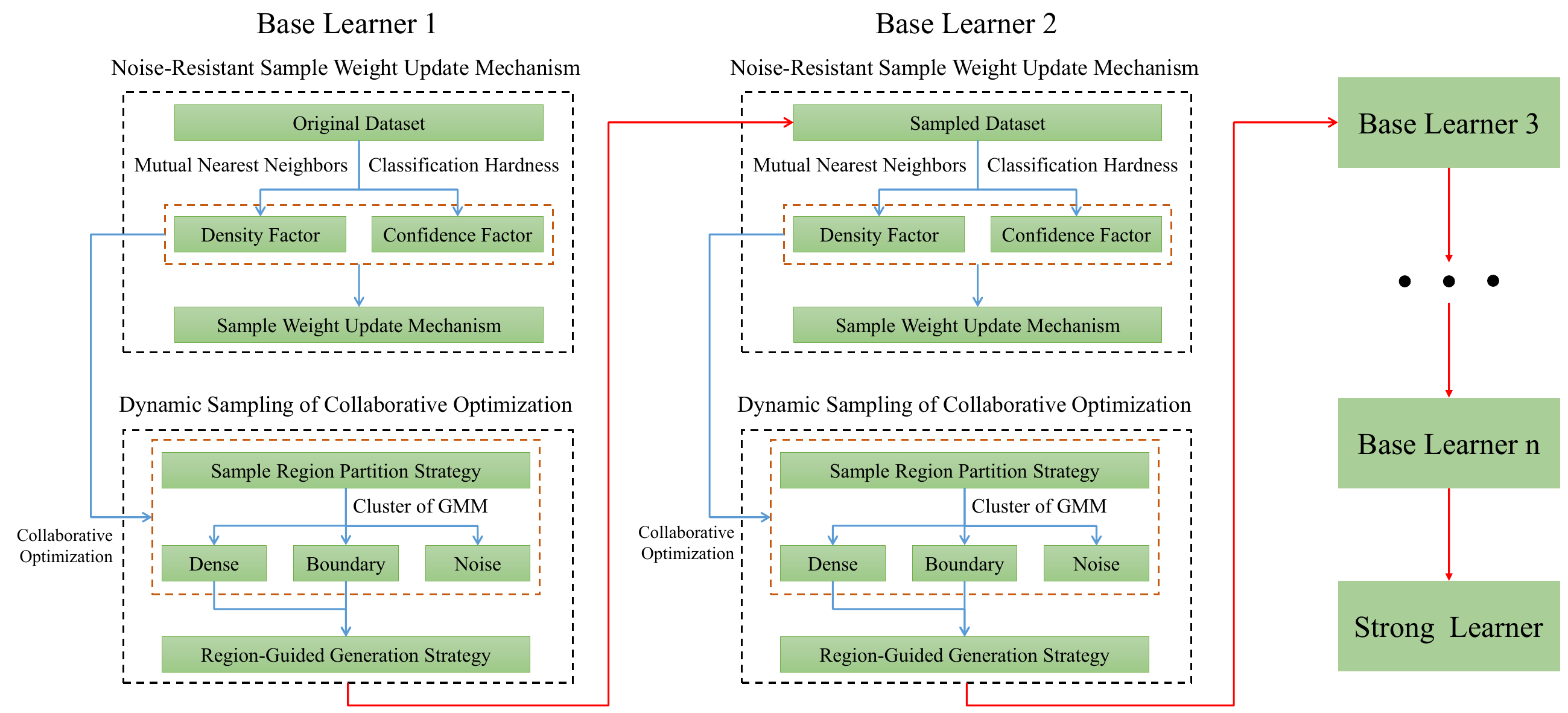}
    \caption{Overall framework of the model proposed in this study.}
    \label{fig:Fig1} 
\end{figure}

The key novelty lies in the closed-loop collaboration between the weight update scheme and the region-guided resampling method, rather than in either component alone. Specifically, the density and confidence factors serve as a shared mechanism that modulates sample weights and etermines the number of synthetic samples generated in dense, boundary, and noise regions. The resampled data, in turn, reshape the ensemble margins and guide weight updates in the next Boosting epoch. Figure \ref{fig:Fig1} illustrates the overall framework, and Algorithm \ref{alg:alg1} shows the pseudocode.

\subsection{Classic AdaBoost Model}
\label{sec:Classic AdaBoost model}

AdaBoost is one of the most prevalent ensemble learning models. It constructs a strong learner by iteratively adjusting sample weights and model weights to integrate multiple base learners. This approach demonstrates superior flexibility and generalization capabilities in both binary and multiclass classification tasks \cite{28}. Given a dataset $D=\{(x_i,y_i)\}_{i=1}^N$ containing $N$ samples, where $x_i\in\mathbb{R}^d$ represents a $d$-dimensional feature vector and $y_i\in\{1,\cdots,c\}$ denotes the corresponding class label. AdaBoost trains $m$ base learners $\{h_t(\cdot)\}_{t=1}^m$ over $m$ iterations. Each base learner has a corresponding model weight $\beta_t$ used to construct the strong learner. For multiclass classification, we adopt the SAMME framework, which obtains the final prediction through weighted majority voting over all base learners.

\begin{algorithm}[H]
    \SetAlgoLined
    \DontPrintSemicolon
    \linespread{1}\selectfont
    
    \caption{Dynamic Sampling through Collaborative Optimization}
    \label{alg:alg1}
    
    \KwIn{Minority class dataset $D_c = \{(x_i, y_i)\}$, Current epoch $t$, Total epochs $m$, Density threshold $\rho_0$}
    \KwOut{Sampled dataset $D'_c$}
    \vspace{0.3em}
    
    \textbf{Stage 1: Adaptive Clustering}\;
    Determine optimal cluster count $g$ via Bayesian Information Criterion\;
    Decompose $D_c$ into $g$ clusters $\{C_1, \dots, C_g\}$ using Gaussian Mixture Model\;
    
    \textbf{Stage 2: Sample Region Partitioning}\;
    \For{each cluster $s \in \{1, \dots, g\}$}{
        \For{each sample $x_i \in C_s$}{
            \uIf{$\rho_i \geq \rho_0$}{
                Assign $x_i \to $ Dense Region\;
            }
            \uElseIf{$\rho_i < \rho_0$ \textbf{and} $H_i > \mu_H - \sigma_H$}{
                Assign $x_i \to $ Boundary Region\;
            }
            \uElse{
                Assign $x_i \to $ Noise Region\;
            }
        }
    }

    \textbf{Stage 3: Dynamic Sampling Allocation}\;
    Compute the base learner target sampling weight $O_t$ $\triangleright$ Eq.(\ref{eq:10})\;
    \For{each cluster $s \in \{1, \dots, g\}$}{
        Compute the target sampling weight $P_s$ $\triangleright$ Eq.(\ref{eq:9})\;
        Calculate the total target sample quantity $T_{st}$ $\triangleright$ Eq.(\ref{eq:8})\;
    }
    
    \textbf{Stage 4: Region-Guided Generation}\;
    Initialize synthetic set $x_{syn} = \emptyset$\;
    \For{each cluster $s \in \{1, \dots, g\}$}{
        \While{$|x_{syn}| < \text{allocated count for } s$}{
            Select parent $x_a$ from Boundary Region and $x_b$ from Dense Region\;
            Generate $x_{new}$ $\triangleright$ Eq.(\ref{eq:11})\;
            Add $x_{new}$ to $x_{syn}$\;
        }
    }
    $D'_c \leftarrow D_c \cup x_{syn}$\;
    
    \Return{$D'_c$}
\end{algorithm}

\subsection{Noise-Resistant Sample Weight Update Mechanism}
\label{sec:Noise-resistant sample weight update mechanism}

Traditional AdaBoost prioritizes hard samples by increasing the weights of misclassified samples. However, because the model evaluates sample importance solely based on the classification error rate, noise samples that are difficult to classify correctly receive progressively larger weights throughout the training process. This limitation is amplified when processing multiclass imbalanced datasets. Traditional AdaBoost forces the model to fit these isolated points, as noise often mimics hard samples with high errors, leading to overfitting and distorted decision boundaries that encroach on minority class regions.

To mitigate the risk of noise-induced overfitting and ensure the model remains focused on informative data, this study proposes a noise-resistant sample weight update mechanism. By integrating both a density factor and a confidence factor, this mechanism prioritizes samples in dense regions and near decision boundaries while penalizing noise.

\subsubsection{Density Factor based Mutual Nearest Neighbors}

We introduce a density factor based on mutual nearest neighbors to quantify local geometric density. For sample $x_i$, if $x_j$ is in its $k$-nearest neighbor set, and $x_i$ is also in the $k$-nearest neighbor set of $x_j$, they constitute mutual nearest neighbors. The mutual nearest neighbor set $\mathcal{N}(x_i)$ of sample $x_i$ is defined as follows:

\begin{equation}
    \mathcal{N}(x_i)={x_j \mid x_j \in \mathcal{N}_k(x_i) \wedge x_i \in \mathcal{N}_k(x_j)},
\label{eq:1}
\end{equation}

\noindent where $\mathcal{N}_{k}(x_{i})$ and $\mathcal{N}_{k}(x_{j})$ represent the $k$-nearest neighbor sets of samples $x_i$ and $x_j$, respectively.

Next, we define the density factor $\rho_i$ of sample $x_i$. The factor $\rho_i$ quantifies the density of the region where the sample is located through the cardinality of mutual nearest neighbors. It is normalized to the range $[0, 1]$:

\begin{equation}
\rho_i = \frac{|\mathcal{N}(x_i)| - \min_j |\mathcal{N}(x_j)|}{\max_j |\mathcal{N}(x_j)| - \min_j |\mathcal{N}(x_j)|},
\label{eq:2}
\end{equation}

\noindent where $|\mathcal{N}(x_i)|$ and $|\mathcal{N}(x_j)|$ represent the cardinalities of the mutual nearest neighbor sets of samples $x_i$ and $x_j$, respectively. When $\rho_i$ approaches 1, the sample is located in a core high-density region. When $\rho_i$ approaches 0, the sample is located in a sparse or isolated region.

In this study, the density factor is used for both updating sample weights and partitioning sample regions in the dynamic sampling module. We improve the traditional $k$-nearest neighbors by employing mutual nearest neighbors to mitigate density calculation errors triggered by noise. Traditional $k$-nearest neighbors establishes a unidirectional relationship based on distance. This leads to erroneous connections with distant samples. Moreover, the asymmetry of traditional  $k$-nearest neighbors allows noise samples to skew the density evaluation of legitimate samples \cite{29}. 

\begin{figure}
    \centering
    \includegraphics[width=1.0\linewidth]{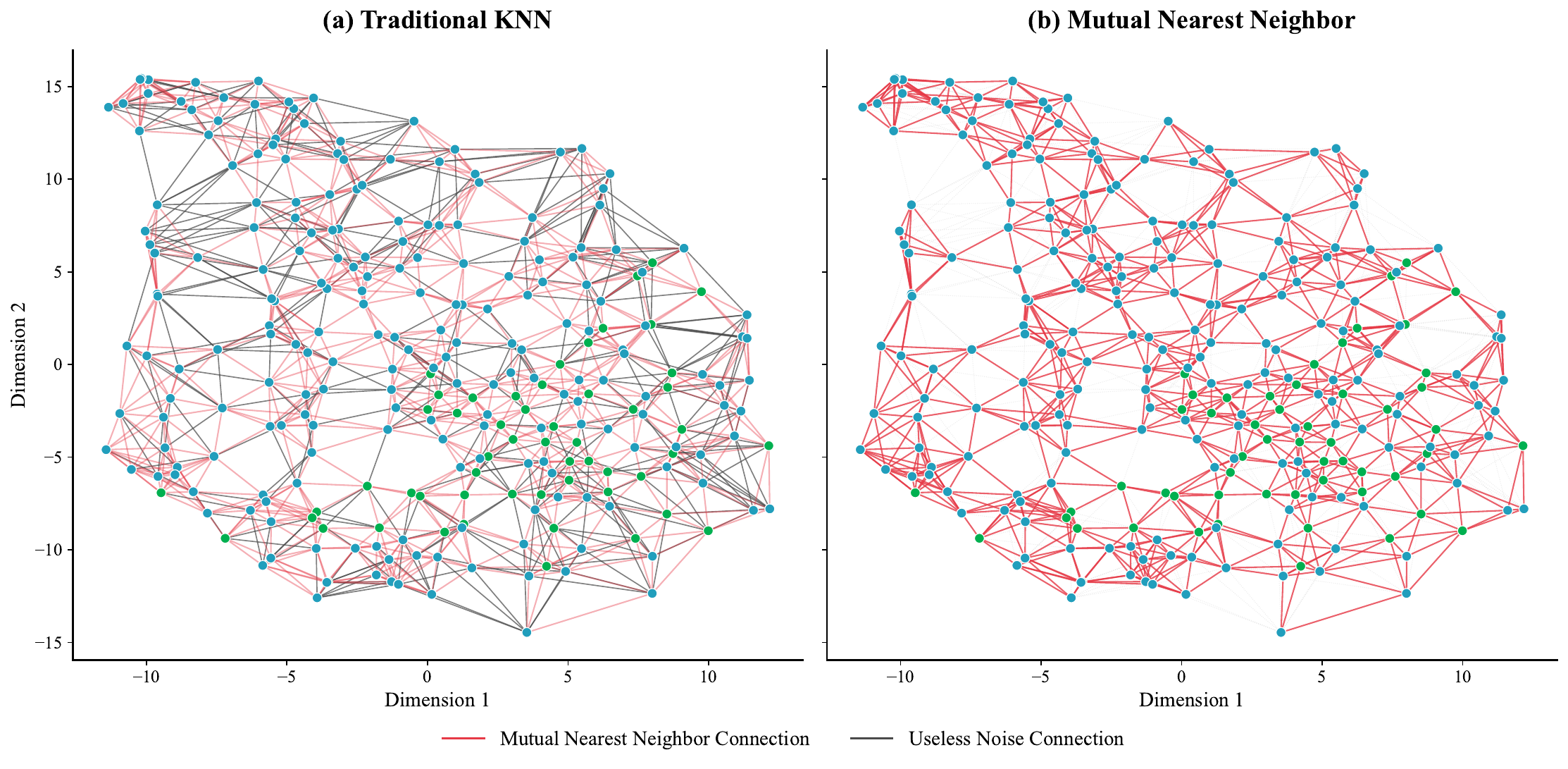}
    \caption{Comparison of connectivity between $k$-nearest neighbors and mutual nearest neighbors.}
    \label{fig:Fig2} 
\end{figure}

Figure \ref{fig:Fig2} visualizes the connection differences between traditional $k$-nearest neighbors and mutual nearest neighbors with $k=10$. The $k$-nearest neighbors establishes 2,670 edges, and the mutual nearest neighbors establishes only 1,141 edges. The significant reduction in the number of edges demonstrates that mutual nearest neighbors effectively filters out noise connections and clarifies the local geometric structure of the data.

\subsubsection{Confidence Factor based Classification Hardness}

The density factor effectively quantifies local sparsity geometrically, it allows the model to distinguish between high-density dense samples and low-density noise. However, density alone cannot distinguish between destructive noise and valuable boundary samples.

To address this, we introduce a confidence factor. This factor measures classification hardness based on the prediction probabilities of the current base learner. We define the classification hardness $H_i$ of training sample $x_i$ as:

\begin{equation}
H_i = \frac{1 - p_{y_i}(x_i) + \max_{j \neq y_i} p_j(x_i)}{2},
\label{eq:3}
\end{equation}

\noindent where $p_{y_i}(x_i)$ represents the prediction probability of sample $x_i$ on the true class $y_i$, and $\max_{j \neq y_i} p_j(x_i)$ represents the model's highest prediction probability for its non-true classes.

The classification hardness $H_{i}$ ranges from 0 to 1. For samples that are easily classified correctly, $p_{y_{i}}(x_{i})$ approaches 1, causing $H_{i}$ to approach 0. For difficult ones, $H_{i}$ approaches 1. A value of $H_{i}$ near 0.5 indicates ambiguity, where the samples typically lie near the decision boundaries.

To utilize hardness for weight adjustment, we map $H_{i}$ to a confidence factor $\delta_{i}$ using a Gaussian function:

\begin{equation}
\delta_i = 1 - \exp\left(-\frac{(H_i - \mu_H)^2}{2\sigma_H^2}\right),
\label{eq:4}
\end{equation}

\noindent where $\mu_H$ and $\sigma_H$ represent the mean and standard deviation of classification hardness across all training samples.

When a sample's hardness approaches the mean, $\delta_i$ approaches 0, indicating minimum model confidence. This characterizes ambiguous boundary samples. Conversely, when $\delta_i$ approaches 1, the sample is either extremely easy or extremely hard to classify. These extremes generally correspond to simple samples or noise, respectively.

\subsubsection{Sample Weight Update Formula}

The density factor and confidence factor characterize samples from two complementary perspectives: geometric structure and classification hardness. Relying on a single factor is insufficient. Specifically, the density factor identifies sparse samples but cannot determine whether the sparsity stems from meaningless noise or valuable decision, because they are all difficult to classify correctly. Conversely, relying solely on the confidence factor mirrors traditional AdaBoost, emphasizing hard samples excessively and often leading to overfitting on noise.

The cooperative action of these two factors is key to achieving noise resistance and imbalanced learning. The density factor partitions samples into dense and non-dense regions. To distinguish between isolated noise samples and ambiguous boundary samples, the non-dense regions are further divided into boundary and noise regions using the confidence factor. 

We fuse the density factor and confidence factor to construct a novel sample weight update mechanism. Given the sample weight $w^{(t)}$ in the $t$-th epoch, the updated weight $w^{(t+1)}$ is defined as:

\begin{equation}
\omega^{(t+1)} = \exp\left(-\delta^{(t)} \cdot (1 - \rho^{(t)})\right) \cdot \omega^{(t)} \cdot \exp\left(-\beta^{(t)} \cdot \mathbb{I}{[h_t(x_i) = y_i]}\right),
\label{eq:5}
\end{equation}

\noindent where $\rho^{(t)}$ and $\delta^{(t)}$ represent the density factor and confidence factor of the $t$-th epoch, respectively. $\mathbb{I}_{[h_t(x_i)=y_i]}$ represents the indicator function, which is 1 when the prediction of the base learner $h_t$ for sample $x_i$ is correct, and 0 otherwise. $\beta^{(t)}$ is the voting weight of AdaBoost. Its calculation follows the concept of traditional AdaBoost, the lower the classification error rate of the base learner, the larger its weight:

\begin{equation}
\beta^{(t)} = \frac{1}{2} \ln\left(\frac{1 - \epsilon_t}{\epsilon_t}\right) = \frac{1}{2} \ln\left(\frac{\sum_{h_t(x_i) = y_i} \omega^{(t)}(x_i, y_i)}{\sum_{h_t(x_i) \neq y_i} \omega^{(t)}(x_i, y_i)}\right),
\label{eq:6}
\end{equation}

\noindent where $\epsilon_t$ represents the classification error rate of the $t$-th epoch base learner. 

\begin{figure}
    \centering
    \includegraphics[width=1.0\linewidth]{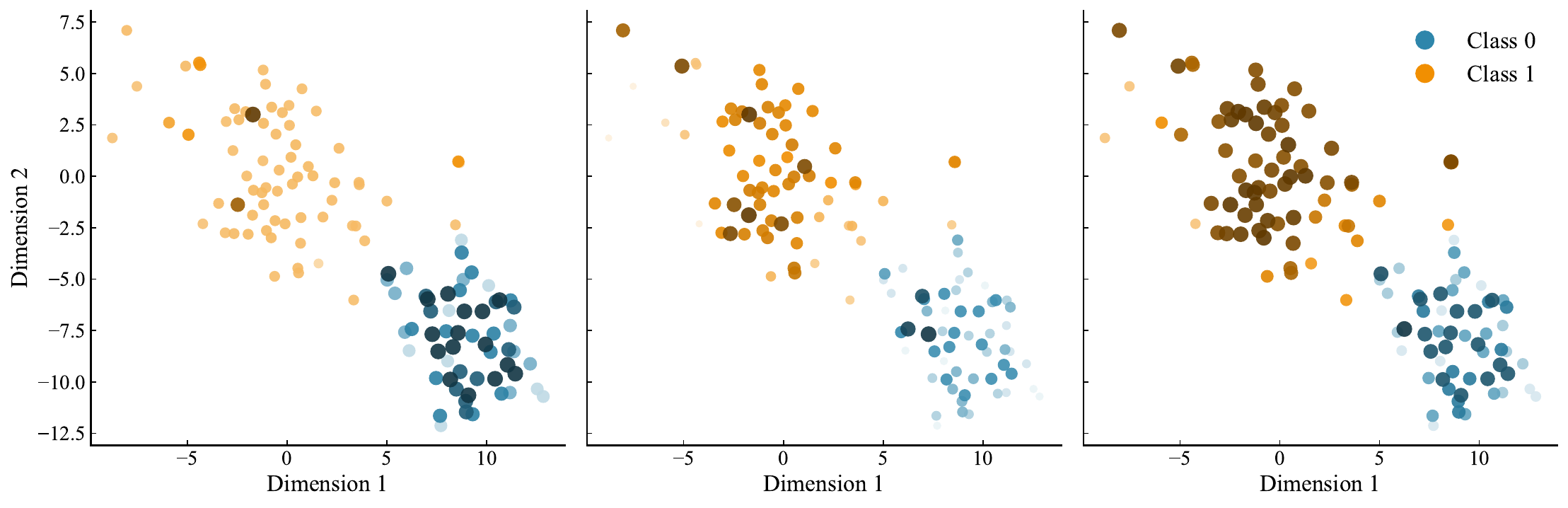}
    \caption{Evolution of sample weights during training process, darker shades indicate higher weights.}
    \label{fig:Fig3} 
\end{figure}

Figure \ref{fig:Fig3} visualizes the changes in sample weights as the number of iterations increases. Samples located in dense regions receive increased weights because they provide valuable structural information for model fitting. Conversely, the mechanism reduces the weights of noisy samples situated far from the cluster centers. Meanwhile, boundary samples receive progressively greater attention across iterations to reconstruct clear decision boundaries. This targeted weighting strategy ensures that the model focuses on high-value samples while remaining robust to noise. 

The pseudocode for the improved sample weight update mechanism is shown in Algorithm \ref{alg:alg2}.

\begin{algorithm}[H]
    \SetAlgoLined
    \linespread{1}\selectfont
    \caption{Noise-Resistant Sample Weight Update Mechanism}
    \label{alg:alg2} 

    \KwIn{Dataset $D=\{(x_i, y_i)\}_{i=1}^N$, Base learner $h_t$, Sample weights $\omega^{(t)}$, Neighbor count $k$}
    \KwOut{Updated sample weights $\omega^{(t+1)}$}
    
    \textbf{Stage 1: Compute Density Factor}\;
    \For{$i = 1$ \KwTo $N$}{
        Identify mutual nearest neighbor set $\mathcal{N}(x_i)$ $\triangleright$ Eq.(\ref{eq:1})\;
        Compute density factor $\rho_i^{(t)}$ based on mutual nearest neighbor $\triangleright$ Eq.(\ref{eq:2})\;
    }

    \textbf{Stage 2: Compute Confidence Factor}\;
    Initialize hardness vector $\mathbf{H} = [H_1, \dots, H_N]$\;
    \For{$i = 1$ \KwTo $N$}{
        Obtain prediction probabilities $p(x_i)$ from $h_t$\;
        Compute classification hardness $H_i$ based on  prediction probabilities $\triangleright$ Eq.(\ref{eq:3})\;
    }
    Compute mean $\mu_H$ and standard deviation $\sigma_H$ of vector $\mathbf{H}$\;
    \For{$i = 1$ \KwTo $N$}{
        Compute confidence factor $\delta_i^{(t)}$ using Gaussian mapping $\triangleright$ Eq.(\ref{eq:4})\;
    }
        
    \textbf{Stage 3: Update Sample Weights}\;
    Compute base learner error rate $\epsilon_t$\;
    Compute voting weight $\beta^{(t)}$ $\triangleright$ Eq.(\ref{eq:6})\;
    \For{$i = 1$ \KwTo $N$}{
        Update weight $\omega_i^{(t+1)}$ by fusing $\rho_i^{(t)}$, $\delta_i^{(t)}$, and $\beta^{(t)}$  $\triangleright$ Eq.(\ref{eq:5})\;
    }
    Normalize weights $\omega^{(t+1)}$ such that $\sum_{i=1}^N \omega_i^{(t+1)} = 1$\;
    
    \Return{$\omega^{(t+1)}$}
\end{algorithm}

\subsection{Dynamic Sampling through Collaborative Optimization}
\label{sec:Dynamic sampling through collaborative optimization}

Traditional algorithm-level methodologies typically combine data sampling and model training in a simple manner, isolating the sampling process from the ensemble learning model and limiting the imbalanced learning performance. To address this issue, this study designs a dynamic sampling strategy based on the density factor and confidence factor.

Furthermore, to overcome the drawbacks of blind interpolation in traditional oversampling methods, we design a region-guided generation strategy. This strategy requires that one parent sample originates from the boundary region and the other from the dense region, drawing ambiguous boundary samples toward the dense region. In addition, we propose a progressive sampling scheduler to manage the synthesis process. This scheduler adaptively controls the number of synthetic samples generated for each base learner and each cluster by incorporating both the density factor and the confidence factor.

\subsubsection{Sample Region Partitioning Strategy}

Traditional sampling methods typically perform random sampling or undersampling across the entire sample space. These methods often ignore local distribution heterogeneity, which can degrade the quality of generated samples or reduce the effectiveness of sample elimination. To address this, we first partition the minority class into multiple clusters. Sample region partitioning and synthesis are then performed independently within each cluster.

When oversampling class $c$, it is treated as the minority class. All other classes with larger sample sizes than class $c$ are categorized as the majority class. Within each cluster $s$, the rules for sample region partitioning are defined as follows:

\begin{equation}
\{(x_i,y_i)\}_s\in\begin{cases}
\mathcal{S}_1,&\rho_i\geq\rho_0,\\
\mathcal{S}_2,&\rho_i<\rho_0\quad\text{and}\quad H_i>\mu_H-\sigma_H,\\
\mathcal{S}_3,&\rho_i<\rho_0\quad \text{and}\quad H_i\le\mu_H-\sigma_H,
\end{cases}
\quad s=1,\cdots,g,
\label{eq:7}
\end{equation}

\noindent where $\rho_i$ denotes the density factor of the $i$-th sample within cluster $s$, $g$ represents the total number of clusters, which is determined using the Bayesian Information Criterion. Additionally, $\rho_0$ serves as a predefined density threshold, and $\mu_H$ and $\sigma_H$ represent the mean and standard deviation of sample confidence, respectively. Based on these parameters, the equation partitions the samples within each cluster into the following three regions:

\begin{enumerate}[label=(\arabic*), leftmargin=*]
\item \textbf{Dense Region} $\mathcal{S}_1$: High density and far from the cluster boundaries. These samples are typically easy to classify.
\item \textbf{Boundary Region} $\mathcal{S}_2$: Lower density but higher confidence. These samples are typically located near the cluster boundaries and easily confused with other classes.
\item \textbf{Noise Region} $\mathcal{S}_3$: Both density and confidence are lower. These samples are typically far from the cluster center and may represent noise far from the decision boundaries.
\end{enumerate}

This clustering-based partitioning strategy ensures targeted sampling operations. It provides a reliable guarantee for generating high-quality samples that maintain similar distributions, thereby clarifying decision boundaries. Furthermore, using the mean and standard deviation of the confidence factor as criteria in Eq.(\ref{eq:7}) reduces the need for manual threshold tuning. This setting also enables the adaptive determination of parameters based on the specific distribution of the dataset.

\subsubsection{Region-Guided Synthetic Sample Generation}

Traditional sampling methods typically balance datasets by increasing the minority class count to match the majority class in a single step. In contrast, we propose a progressive sampling scheduler that allocates a specific sampling quantity to each training epoch. 

The progressive sampling scheduler integrates dynamic sampling directly into the iterative training process. Specifically, it first determines the total target sampling quantity for each class, and then dynamically assigns sampling weights to different clusters across epochs. These weights are adaptively determined based on both cluster-level and base-learner target sampling requirements.

For the currently processed class $c$ in the $t$-th epoch, the target sampling quantity $T_{st}$ of cluster $s$ is calculated as:

\begin{equation}
T_{st} = \lfloor P_s \cdot O_t \cdot (\max_{c'} N(c') - N(c)) \rfloor, \quad s = 1, \dots, g, \quad t = 1, \dots, m,
\label{eq:8}
\end{equation}

\noindent where $\max_{c'} N(c') - N(c)$ represents the total target sample quantity required for the class $c$, $\lfloor \cdot \rfloor$ represents the floor function, $c'$ represents the majority class, $O_t$ represents the target sampling weight of the $t$-th base learner. In addition, $P_s$ represents the target sampling weight of cluster $s$, it is calculated as follows:

\begin{equation}
P_s = \frac{\rho_s^{avg} \cdot |N_s|}{\sum_{s=1}^{g} \rho_s^{avg} \cdot |N_s|},
\label{eq:9}
\end{equation}

\noindent where $\rho_s^{avg}$ represents the average density factor of the $s$-th cluster, and $|N_s|$ represents the sample quantity of the $s$-th cluster.

The base learner target sampling weight $O_t$ is calculated as:

\begin{equation}
O_t = \frac{\tan\left(\frac{m-t}{m-1} \cdot \frac{\pi}{4}\right)}{\sum_{t=1}^{m} \tan\left(\frac{m-t}{m-1} \cdot \frac{\pi}{4}\right)},
\label{eq:10}
\end{equation}

\noindent where $m$ is the total number of AdaBoost iterations.

The cluster target sampling weight $P_s$ accounts for both sample density and cluster size, it assigns higher weights to clusters with high density and large sample counts, as these clusters typically contain high-quality samples and provide higher fitting value.

Meanwhile, the base learner target sampling weight $O_t$ is determined by a tangent function. This function allocates larger sampling weights during the early stages of training. As training progresses, the weight gradually decreases. This strategy allows the model to quickly compensate for the minority class deficit initially. As the decision boundaries stabilize, the sampling intensity decreases, ensuring a smoother and more effective learning process.

Finally, after determining the target sampling quantity $T_{st}$ for cluster $s$ in the $t$-th epoch, we perform region-guided sample synthesis as follows:

\begin{equation}
x_{new} = x_a + \lambda (x_b - x_a), \quad x_a \in \mathcal{S}_2, \quad x_b \in \mathcal{S}_1,\quad \lambda \in \mathcal{U}(0, 1)
\label{eq:11}
\end{equation}

\noindent where $x_a$ and $x_b$ represent parent samples from the boundary region and dense region, respectively.

\begin{figure}
    \centering
    \includegraphics[width=0.9\linewidth]{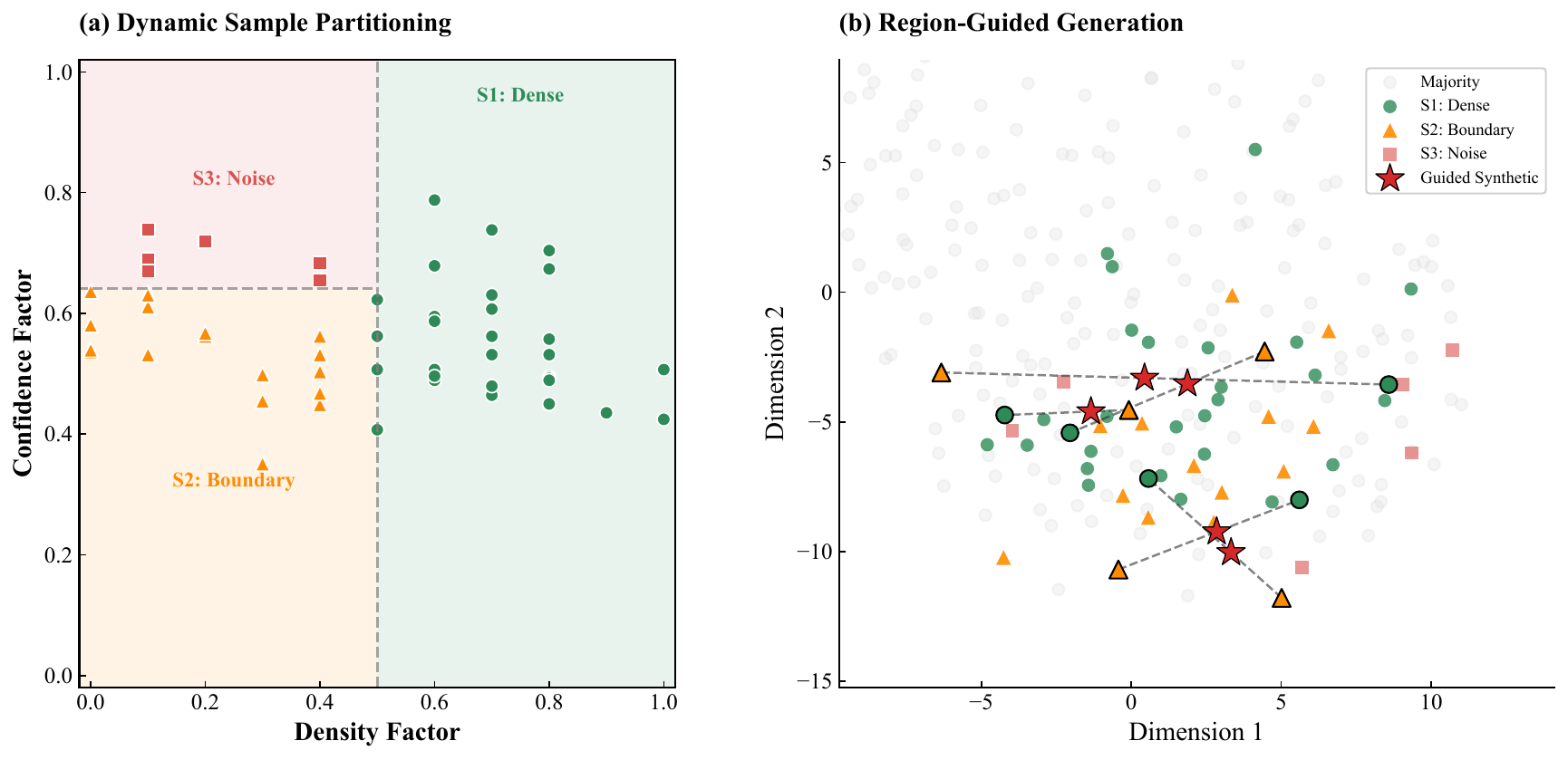}
    \caption{Schematic diagram of dynamic sampling under collaborative optimization.}
    \label{fig:Fig4} 
\end{figure}

Figure \ref{fig:Fig4} illustrates the dynamic sampling strategy under cooperative optimization. Specifically, Figure \ref{fig:Fig4}(a) details the sample region partitioning strategy, which categorizes samples into three distinct regions based on their density factor and confidence factor. Figure \ref{fig:Fig4}(b) depicts the region-guided sample synthesis strategy. This strategy locks the boundary samples and performs guided interpolation toward dense samples, thereby generating synthetic samples with high confidence.

\begin{figure}[!htbp]
    \centering
    \includegraphics[width=1.0\linewidth]{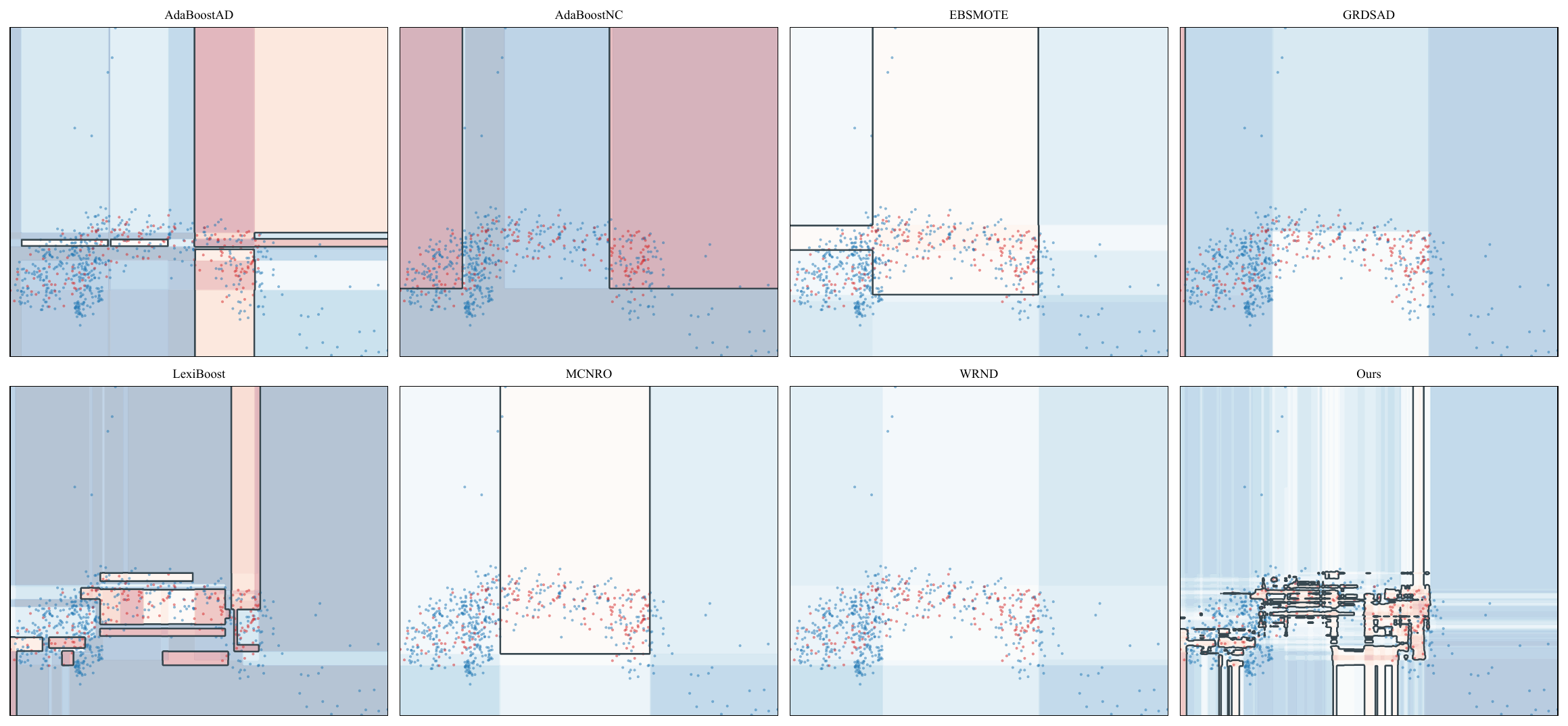}
    \caption{Visualization of decision boundaries between the proposed method and seven comparative models on the vehicle1 dataset.}
    \label{fig:Fig5}
\end{figure}

Figure \ref{fig:Fig5} presents 2D PCA decision-boundary visualizations of different methods on the vehicle1 dataset. The minority class decision regions of WRND and GRDSAD collapse almost completely, making minority samples difficult to distinguish. In contrast, AdaBoostNC excessively expands the minority class region to cover nearly half of the grid, severely encroaching upon the majority class region. LexiBoost exhibits noticeable fragmentation in its decision boundaries. The proposed method preserves the core minority class decision region while avoiding excessive expansion into the minority-class region.

\subsection{Complexity Analysis}
\label{sec:Complexity Analysis}

Let $n$ denote the number of training samples, $d$ the feature dimension, $C$ the number of classes, $T$ the number of Boosting epochs, and $L$ the maximum depth of the base decision tree. Since all Boosting methods train $T$ base learners, their total complexity equals $T$ times the per-epoch complexity.

In addition to training the base learner, the proposed model performs three operations in each epoch. It computes the mutual-nearest-neighbor density with a KD-tree in $O(n\log n)$, computes the confidence factor and updates weights in $O(nC)$, and performs dynamic sampling in $O(n_m d^2 + n_s d)$, where $n_m \le n$ denotes the number of minority samples and $n_s$ denotes the number of generated samples. Training a decision tree of depth $L$ costs $O(ndL)$.

For the algorithm-level baselines in Table \ref{tab:table2}, AdaBoostNC trains a base learner and applies a negative-correlation penalty to the outputs, yielding $O(Tn(dL+C))$. AdaBoostAD estimates local density with $k$-nearest neighbors and updates weights according to the imbalance ratio and density, yielding $O(Tn(k+C+dL))$. GRDSAD builds a $k$-nearest-neighbor graph, performs an LPP projection with eigendecomposition, and trains a base learner, yielding $O(T(n\log n + d^3 + ndL))$. LexiBoost solves a lexicographic linear program to obtain sample weights before training the base learner; the LP-based weight computation dominates, yielding $O(T(n^2 + ndL))$ in the worst case.

\section{Experiments and Discussion}
\label{sec:Experiments and Discussion}

\subsection{Experimental Setting}
\label{sec:Experimental Setting}

In this study, Python 3.12 was used as the programming language for the implementation of the code, with all necessary dependencies installed through the Anaconda distribution, including scikit-learn (version 1.7.1) for the implementation of the machine learning algorithms, Pandas (version 2.3.3) for the data processing, and numerical operations with NumPy (version 2.4.6). All experimental evaluations were conducted on a Windows 11 (64-bit) platform equipped with an Intel Core i9-14900HX processor and 16GB RAM.

\subsection{Dataset}
\label{sec:Dataset}

To ensure statistical significance and generalization, we selected 40 classification datasets from the KEEL repository \cite{32}. These datasets vary widely in their imbalance ratios, sample size, and class counts, covering diverse scenarios, ranging from low to high dimensionality and from mild to extreme imbalance. 

Table \ref{tab:table1} summarizes the characteristics of these datasets, all directly downloaded from the KEEL repository \cite{32}. We used the 5-fold cross-validation versions provided by KEEL for experimental evaluation. These datasets contain no missing values, and KEEL has encoded all categorical attributes. We performed no additional preprocessing. In Table \ref{tab:table1}, \#Samp, \#Feat, \#Class, and IR denote the total numbers of samples, features, and classes, and the ratio of the majority-class size to the minority-class size, respectively.

\begin{table}[!htbp]
\caption{The statistical information of the 40 KEEL datasets used in this study.}
\label{tab:table1}
\small
\resizebox{\textwidth}{!}{%
\begin{tabular}{@{}cccccccccc@{}}
\toprule
Dataset & \#Samp & \#Feat & \#Class & IR & Dataset & \#Samp & \#Feat & \#Class & IR \\
\midrule
abalone-17\_vs\_7-8-9-10 & 2338 & 8 & 2 & 39.31 & kddcup-land\_vs\_satan & 1610 & 41 & 2 & 75.67 \\
abalone-20\_vs\_8-9-10 & 1916 & 8 & 2 & 72.69 & kr-vs-k-three\_vs\_eleven & 2935 & 6 & 2 & 35.23 \\
abalone-21\_vs\_8 & 581 & 8 & 2 & 40.50 & kr-vs-k-zero-one\_vs\_draw & 2901 & 6 & 2 & 26.63 \\
abalone9-18 & 731 & 8 & 2 & 16.40 & kr-vs-k-zero\_vs\_eight & 1460 & 6 & 2 & 53.07 \\
autos & 159 & 25 & 6 & 16.00 & pageblocks & 548 & 10 & 5 & 164.00 \\
car-good & 1728 & 6 & 2 & 24.04 & poker-8-9\_vs\_6 & 1485 & 10 & 2 & 58.40 \\
car-vgood & 1728 & 6 & 2 & 25.58 & poker-8\_vs\_6 & 1477 & 10 & 2 & 85.88 \\
cleveland-0\_vs\_4 & 173 & 13 & 2 & 12.31 & shuttle & 2175 & 9 & 5 & 853.00 \\
dermatology & 358 & 34 & 6 & 5.55 & vehicle1 & 846 & 18 & 2 & 2.90 \\
ecoli-0-1-4-7\_vs\_5-6 & 332 & 6 & 2 & 12.28 & vowel0 & 988 & 13 & 2 & 9.98 \\
ecoli-0-2-6-7\_vs\_3-5 & 224 & 7 & 2 & 9.18 & winequality-red-4 & 1599 & 11 & 2 & 29.17 \\
ecoli-0-3-4-7\_vs\_5-6 & 257 & 7 & 2 & 9.28 & winequality-red-8\_vs\_6 & 656 & 11 & 2 & 35.44 \\
ecoli-0-6-7\_vs\_5 & 220 & 6 & 2 & 10.00 & winequality-red-8\_vs\_6-7 & 855 & 11 & 2 & 46.50 \\
glass & 214 & 9 & 6 & 8.44 & yeast & 1484 & 8 & 10 & 92.60 \\
glass-0-1-4-6\_vs\_2 & 205 & 9 & 2 & 11.06 & yeast-0-2-5-7-9\_vs\_3-6-8 & 1004 & 8 & 2 & 9.14 \\
glass-0-1-6\_vs\_2 & 192 & 9 & 2 & 10.29 & yeast-1\_vs\_7 & 459 & 7 & 2 & 14.30 \\
glass-0-6\_vs\_5 & 108 & 9 & 2 & 11.00 & yeast-2\_vs\_4 & 514 & 8 & 2 & 9.08 \\
glass0 & 214 & 9 & 2 & 2.06 & yeast-2\_vs\_8 & 482 & 8 & 2 & 23.10 \\
glass4 & 214 & 9 & 2 & 15.46 & yeast1 & 1484 & 8 & 2 & 2.46 \\
glass6 & 214 & 9 & 2 & 6.38 & zoo-3 & 101 & 16 & 2 & 19.20 \\
\bottomrule
\end{tabular}}
\end{table}

\subsection{Evaluation Metrics}
\label{sec:Evaluation Metrics}

To evaluate multiclass imbalanced data, we use three macro-averaged metrics: Macro-F1, G-Mean, and AUPRC. Each metric is computed for every class $c\in C$ and then averaged over all classes.

Macro-F1 is the average of the per class F1 scores:

\begin{equation}
\text{Macro-F1}=\frac{1}{|C|}\sum_{c\in C} F1_c,
\end{equation}

\begin{equation}
F1_c=\frac{2P_cR_c}{P_c+R_c},\quad P_c=\frac{TP_c}{TP_c+FP_c},\quad R_c=\frac{TP_c}{TP_c+FN_c},
\end{equation}

\noindent where $TP_c$, $FP_c$, and $FN_c$ denote the numbers of true positives, false positives, and false negatives for class $c$. Macro-F1 reflects balanced performance across classes.

G-Mean is the geometric mean of the per class recalls:

\begin{equation}
\text{G-Mean}=\left(\prod_{c\in C} R_c\right)^{1/|C|}
= \exp\!\left(\frac{1}{|C|}\sum_{c\in C}\ln R_c\right).
\end{equation}

It is zero if any class has zero recall, so it penalizes poor minority class performance.

AUPRC is the macro average of the per class areas under the precision-recall curves. For each class, we use a one-vs-rest setting and rank samples by the predicted probability $\hat{y}_c(x)$:

\begin{equation}
AP_c=\sum_n (R_n-R_{n-1})\,P_n,
\end{equation}

\noindent where $P_n$ and $R_n$ are the precision and recall accumulated up to the $n$-th sample. The final multi class metric is

\begin{equation}
\text{AUPRC}=\frac{1}{|C|}\sum_{c\in C} AP_c.
\end{equation}

\noindent where $P_n$ and $R_n$ are the precision and recall at the $n$-th sample. AUPRC is threshold-independent and is more informative than accuracy or AUC under class imbalance.

\subsection{Baselines}
\label{sec:Baselines}

We evaluated the proposed model against 7 state-of-the-art imbalanced learning methods. These baselines include 4 algorithm-level models and 3 data-level models. To ensure a fair comparison, all data-level models used AdaBoost as the base classifier, and all models used 30 base learners. We tuned hyperparameters with Bayesian optimization implemented in Optuna over 30 iterations, using the average of Macro-F1, G-Mean, and AUPRC as the objective.

\begin{table}[!htbp]
\caption{Overview of the seven baseline methods compared in this study.}
\label{tab:table2}
\centering
\small
\begin{tabular}{@{}ccc@{}}
\toprule
Model & Category & Brief Description \\
\midrule
AdaBoostAD & Algorithm-level & Weighting mechanism based on imbalance ratios and local density \\
AdaBoostNC & Algorithm-level & Negative correlation penalty to enhance base classifier diversity \\
EB-SMOTE & Data-level & Expands minority and majority borderlines toward each other \\
GRDSAD & Algorithm-level & Graph discriminative dynamic sampling AdaBoost \\
LexiBoost & Algorithm-level & Lexicographic optimization for inter class trade-off balancing \\
MCNRO & Data-level & Neighborhood re-partitioning for multiclass scenarios \\
WRND & Data-level & Weighted oversampling with relative neighborhood density \\
\bottomrule
\end{tabular}
\end{table}

Table \ref{tab:table2} provides a detailed summary of these baseline models, and Table \ref{tab:table3} lists the hyperparameter search ranges. The optimal hyperparameter combinations identified in this study, along with the associated code and datasets, are available in the GitHub repository: \href{https://github.com/ChuantaoLi/DARG}{https://github.com/ChuantaoLi/DARG}. We employed Macro-F1 score, G-Mean, and AUPRC as evaluation metrics. These metrics provide a comprehensive assessment of model performance on multiclass imbalanced datasets.

\begin{table}[!htbp]
\caption{Hyperparameter search spaces used by Bayesian optimization.}
\label{tab:table3}
\centering
\small
\begin{tabularx}{\textwidth}{@{}ccccc>{\centering\arraybackslash}X@{}}
\toprule
Model & Hyperparameter & Type & Search range & Step & Brief description \\
\midrule
\multirow{3}{*}{Ours} & $k$ & int & [2, 20] & -- & MNN count for density estimation \\
 & $d_{\max}$ & int & [1, 50] & -- & Max depth of base decision tree \\
 & $\rho_0$ & float & [0.1, 0.9] & 0.05 & Density threshold separating safe/boundary/noise regions \\
AdaBoostAD & $k$ & int & [1, 20] & -- & Neighborhood size for local density estimation \\
AdaBoostNC & $\lambda$ & float & [0.5, 5.0] & -- & Negative-correlation penalty weight \\
\multirow{2}{*}{EB-SMOTE} & $\alpha$ & int & [1, 10] & -- & Coefficient of borderline expansion \\
 & $\lambda_b$ & float & [1.0, 5.0] & -- & Width of the borderline region \\
\multirow{2}{*}{GRDSAD} & $d_{\max}$ & int & [1, 20] & -- & Max depth of base decision tree \\
 & $k_{\mathrm{LPP}}$ & int & [3, 20] & -- & Neighbor count for LPP-based projection \\
LexiBoost & $d_{\max}$ & int & [1, 10] & -- & Max depth of base decision tree \\
\multirow{2}{*}{MCNRO} & $k_1$ & int & [1, 20] & -- & Inner-neighborhood size for potential energy \\
 & $k_2$ & int & [1, 20] & -- & Outer-neighborhood size for potential energy \\
WRND & $\eta$ & float & [0.1, 0.9] & 0.05 & Noise ratio for noise filtering \\
\bottomrule
\end{tabularx}
\end{table}

\subsection{Performance Comparison}
\label{sec:Performance Comparison}

This section compares the proposed model with seven baselines on 40 datasets. Table \ref{tab:table7} summarizes the average values, average ranks, and statistical test results of all eight methods, while Table \ref{tab:table4} reports detailed Macro-F1, G-Mean, and AUPRC results on each dataset. Each entry in Table \ref{tab:table4} gives the 5-fold mean and standard deviation. Superscripts indicate dataset-level ranks, with smaller values indicating better performance, and boldface marks the best result in each row. The second-to-last row reports the average rank of each method across all datasets, and the last row counts the wins, ties, and losses of each baseline against the proposed model.

The proposed model achieves the best average ranks on all three metrics, with average ranks of 1.41, 1.88, and 2.12 for Macro-F1, G-Mean, and AUPRC, respectively. The Friedman test indicates significant overall differences among the eight methods, with $\chi^2$ values of 143.357, 108.984, and 137.550, respectively, and all p-values below 0.001. The Holm-corrected Wilcoxon signed-rank test further confirms that the proposed model significantly outperforms every baseline on all three metrics at the 0.05 level. For Macro-F1, the proposed model outperforms AdaBoostAD on all 40 datasets and outperforms each baseline on at least 34 datasets.

The proposed model couples noise-resistant weight updates with region-guided dynamic sampling. The density and confidence factors jointly suppress isolated noise and emphasize genuine boundary samples, while region-guided generation fills sparse boundary areas and reconstructs clearer decision boundaries. In contrast, sampling baselines generate synthetic samples from static geometric information without sensing the current ensemble state, and Boosting baselines only adjust sample weights without changing the training distribution. Consequently, the proposed model performs best on multiclass and highly imbalanced tasks with severe class overlap, whereas simpler or more compact tasks leave less room for improvement.

In summary, the proposed model achieves the best and statistically significant performance across all three metrics, with the most evident gains on complex multiclass imbalanced data.

\begin{table}[!htbp]
\caption{Average performance and Holm-corrected Wilcoxon test results.}
\label{tab:table7}
\resizebox{\textwidth}{!}{%
\begin{tabular}{@{}cccccccccc@{}}
\toprule
 & \multicolumn{3}{c}{Macro-F1} & \multicolumn{3}{c}{G-Mean} & \multicolumn{3}{c}{AUPRC} \\
\cmidrule(lr){2-4}\cmidrule(lr){5-7}\cmidrule(lr){8-10}
Model & Mean & Wilcoxon $p$ & Avg.\ rank & Mean & Wilcoxon $p$ & Avg.\ rank & Mean & Wilcoxon $p$ & Avg.\ rank \\
\midrule
Ours & \textbf{0.811} & -- & \textbf{1.41} & \textbf{0.753} & -- & \textbf{1.88} & \textbf{0.800} & -- & \textbf{2.12} \\
AdaBoostAD & 0.511 & 0.000$^{*}$ & 7.28 & 0.590 & 0.000$^{*}$ & 4.39 & 0.640 & 0.000$^{*}$ & 6.47 \\
AdaBoostNC & 0.570 & 0.000$^{*}$ & 6.22 & 0.454 & 0.000$^{*}$ & 6.21 & 0.618 & 0.000$^{*}$ & 6.38 \\
EB-SMOTE & 0.698 & 0.000$^{*}$ & 3.89 & 0.581 & 0.000$^{*}$ & 4.15 & 0.745 & 0.002$^{*}$ & 3.21 \\
GRDSAD & 0.680 & 0.000$^{*}$ & 4.56 & 0.658 & 0.000$^{*}$ & 3.85 & 0.669 & 0.000$^{*}$ & 5.83 \\
LexiBoost & 0.750 & 0.000$^{*}$ & 3.77 & 0.679 & 0.000$^{*}$ & 3.56 & 0.716 & 0.000$^{*}$ & 5.41 \\
MCNRO & 0.684 & 0.000$^{*}$ & 4.29 & 0.470 & 0.000$^{*}$ & 5.86 & 0.751 & 0.003$^{*}$ & 2.96 \\
WRND & 0.682 & 0.000$^{*}$ & 4.58 & 0.464 & 0.000$^{*}$ & 6.10 & 0.743 & 0.002$^{*}$ & 3.61 \\
\midrule
Friedman $\chi^2$ (p) & \multicolumn{3}{c}{143.357 (1.01e-27)} & \multicolumn{3}{c}{108.984 (1.49e-20)} & \multicolumn{3}{c}{137.550 (1.66e-26)} \\
\bottomrule
\end{tabular}}
\smallskip
{\footnotesize $*$: significant at $\alpha=0.05$ after Holm correction for seven comparisons within each metric.}
\end{table}

\subsection{Ablation Analysis}
\label{sec:Ablation Analysis}

To further verify the contributions of the core components, we conducted an ablation analysis. Table \ref{tab:ablationprog} reports the Macro-F1, G-Mean, and AUPRC results, with boldface marking the best value in each row. We compare five configurations. "Base" denotes standard AdaBoost without any added component. "DF" and "CF" activate only the density factor and confidence factor, respectively. "DF+CF" uses both weight factors, and "Full" further adds region-guided dynamic sampling, denoted as "RG".

Single factors yield limited improvements in Macro-F1 and G-Mean. "DF" reaches 0.758 and 0.634, while "CF" reaches 0.761 and 0.649. For AUPRC, "DF" alone improves the value from 0.726 to 0.762. Combining "DF" and "CF" produces consistent gains on all three metrics, and adding "RG" yields the main improvement, increasing the values to 0.811, 0.753, and 0.800.

The density factor alone suppresses isolated noise but may discard useful boundary samples, whereas the confidence factor alone can overemphasize noise in overlapping regions. Combining them helps distinguish noise from genuinely hard samples and refines the weight update. The decisive gain comes from "RG", which pairs each boundary sample with a dense sample, generates synthetic data along the segment, and removes low-confidence noise, thereby reconstructing clearer decision boundaries. On a few small or compact datasets, synthetic samples may introduce extra noise, so Base or "DF" can perform best; however, Full remains optimal on most datasets.

\subsection{Parameter Sensitivity Analysis}
\label{sec:Parameter Sensitivity Analysis}

We then examine the robustness of the proposed model with respect to its key hyperparameters. Table \ref{tab:senskabs_k} reports sensitivity to the mutual-nearest-neighbor count $k\in\{2, 5, 10, 15, 20\}$, and Table \ref{tab:senskabs_rho} reports sensitivity to the density threshold $\rho_0\in\{0.1, 0.2, 0.3, 0.5, 0.7, 0.9\}$. For each dataset, we vary one parameter at a time, keep the others at their tuned values, and report the five-fold mean and standard deviation.

\begin{sidewaystable}[p]
\caption{Macro-F1, G-Mean, and AUPRC results of all methods on 40 datasets.}
\label{tab:table4}
\small
\resizebox{\textwidth}{!}{%
\begin{tabular}{@{}ccccccccc@{}}
\toprule
Dataset & AdaBoostAD & AdaBoostNC & EB-SMOTE & GRDSAD & LexiBoost & MCNRO & WRND & Ours \\
\midrule
abalone-17\_vs\_7-8-9-10 & 0.402$\pm$0.046\textsuperscript{8} & 0.513$\pm$0.043\textsuperscript{6} & 0.551$\pm$0.057\textsuperscript{5} & 0.479$\pm$0.149\textsuperscript{7} & 0.565$\pm$0.123\textsuperscript{4} & 0.566$\pm$0.048\textsuperscript{2} & 0.566$\pm$0.048\textsuperscript{2} & \textbf{0.669$\pm$0.026\textsuperscript{1}} \\
abalone-20\_vs\_8-9-10 & 0.456$\pm$0.025\textsuperscript{8} & 0.497$\pm$0.025\textsuperscript{6} & 0.636$\pm$0.084\textsuperscript{2} & 0.526$\pm$0.048\textsuperscript{5} & 0.492$\pm$0.028\textsuperscript{7} & 0.558$\pm$0.086\textsuperscript{4} & 0.583$\pm$0.082\textsuperscript{3} & \textbf{0.660$\pm$0.109\textsuperscript{1}} \\
abalone-21\_vs\_8 & 0.575$\pm$0.067\textsuperscript{8} & 0.803$\pm$0.092\textsuperscript{2} & 0.772$\pm$0.077\textsuperscript{7} & 0.802$\pm$0.038\textsuperscript{3} & 0.778$\pm$0.123\textsuperscript{6} & 0.782$\pm$0.081\textsuperscript{4} & 0.782$\pm$0.081\textsuperscript{4} & \textbf{0.853$\pm$0.106\textsuperscript{1}} \\
abalone9-18 & 0.518$\pm$0.127\textsuperscript{8} & 0.553$\pm$0.097\textsuperscript{7} & 0.628$\pm$0.152\textsuperscript{5} & \textbf{0.736$\pm$0.152\textsuperscript{1}} & 0.678$\pm$0.165\textsuperscript{4} & 0.695$\pm$0.147\textsuperscript{3} & 0.628$\pm$0.152\textsuperscript{5} & 0.732$\pm$0.135\textsuperscript{2} \\
autos & 0.152$\pm$0.051\textsuperscript{8} & 0.186$\pm$0.079\textsuperscript{7} & 0.459$\pm$0.146\textsuperscript{4} & 0.539$\pm$0.147\textsuperscript{3} & 0.805$\pm$0.089\textsuperscript{2} & 0.435$\pm$0.111\textsuperscript{6} & 0.452$\pm$0.119\textsuperscript{5} & \textbf{0.841$\pm$0.021\textsuperscript{1}} \\
\bottomrule
\end{tabular}

\begin{tabular}{@{}cccccccc@{}}
\toprule
AdaBoostAD & AdaBoostNC & EB-SMOTE & GRDSAD & LexiBoost & MCNRO & WRND & Ours \\
\midrule
0.697$\pm$0.045\textsuperscript{2} & 0.413$\pm$0.268\textsuperscript{5} & \textbf{0.716$\pm$0.177\textsuperscript{1}} & 0.648$\pm$0.088\textsuperscript{4} & 0.260$\pm$0.373\textsuperscript{6} & 0.257$\pm$0.152\textsuperscript{7} & 0.257$\pm$0.152\textsuperscript{7} & 0.670$\pm$0.046\textsuperscript{3} \\
\textbf{0.807$\pm$0.033\textsuperscript{1}} & 0.437$\pm$0.288\textsuperscript{6} & 0.620$\pm$0.369\textsuperscript{3} & 0.731$\pm$0.409\textsuperscript{2} & 0.619$\pm$0.359\textsuperscript{4} & 0.179$\pm$0.245\textsuperscript{8} & 0.260$\pm$0.238\textsuperscript{7} & 0.561$\pm$0.352\textsuperscript{5} \\
0.821$\pm$0.072\textsuperscript{2} & 0.754$\pm$0.177\textsuperscript{5} & 0.696$\pm$0.119\textsuperscript{8} & 0.801$\pm$0.146\textsuperscript{3} & 0.796$\pm$0.147\textsuperscript{4} & 0.697$\pm$0.119\textsuperscript{6} & 0.697$\pm$0.119\textsuperscript{6} & \textbf{0.884$\pm$0.096\textsuperscript{1}} \\
0.687$\pm$0.136\textsuperscript{3} & 0.382$\pm$0.392\textsuperscript{6} & 0.343$\pm$0.331\textsuperscript{7} & 0.702$\pm$0.216\textsuperscript{2} & 0.588$\pm$0.252\textsuperscript{4} & 0.473$\pm$0.297\textsuperscript{5} & 0.343$\pm$0.331\textsuperscript{7} & \textbf{0.737$\pm$0.188\textsuperscript{1}} \\
0.000$\pm$0.000\textsuperscript{6} & 0.000$\pm$0.000\textsuperscript{6} & 0.122$\pm$0.272\textsuperscript{4} & 0.287$\pm$0.397\textsuperscript{3} & 0.656$\pm$0.372\textsuperscript{2} & 0.112$\pm$0.251\textsuperscript{5} & 0.000$\pm$0.000\textsuperscript{6} & \textbf{0.826$\pm$0.050\textsuperscript{1}} \\
0.744$\pm$0.037\textsuperscript{4} & 0.589$\pm$0.020\textsuperscript{6} & 0.632$\pm$0.066\textsuperscript{5} & 0.813$\pm$0.354\textsuperscript{3} & 0.913$\pm$0.033\textsuperscript{2} & 0.000$\pm$0.000\textsuperscript{7} & 0.000$\pm$0.000\textsuperscript{7} & \textbf{0.939$\pm$0.021\textsuperscript{1}} \\
\bottomrule
\end{tabular}

\begin{tabular}{@{}cccccccc@{}}
\toprule
AdaBoostAD & AdaBoostNC & EB-SMOTE & GRDSAD & LexiBoost & MCNRO & WRND & Ours \\
\midrule
0.539$\pm$0.022\textsuperscript{7} & 0.559$\pm$0.028\textsuperscript{5} & 0.594$\pm$0.073\textsuperscript{4} & 0.539$\pm$0.031\textsuperscript{6} & 0.535$\pm$0.068\textsuperscript{8} & 0.638$\pm$0.019\textsuperscript{2} & 0.630$\pm$0.039\textsuperscript{3} & \textbf{0.666$\pm$0.018\textsuperscript{1}} \\
0.547$\pm$0.008\textsuperscript{5} & 0.519$\pm$0.006\textsuperscript{8} & 0.690$\pm$0.133\textsuperscript{2} & 0.543$\pm$0.019\textsuperscript{6} & 0.531$\pm$0.014\textsuperscript{7} & 0.667$\pm$0.097\textsuperscript{3} & 0.655$\pm$0.099\textsuperscript{4} & \textbf{0.708$\pm$0.119\textsuperscript{1}} \\
0.701$\pm$0.095\textsuperscript{8} & 0.744$\pm$0.102\textsuperscript{6} & 0.800$\pm$0.096\textsuperscript{3} & 0.711$\pm$0.016\textsuperscript{7} & 0.781$\pm$0.074\textsuperscript{5} & 0.808$\pm$0.109\textsuperscript{2} & 0.798$\pm$0.111\textsuperscript{4} & \textbf{0.822$\pm$0.116\textsuperscript{1}} \\
0.652$\pm$0.109\textsuperscript{6} & 0.624$\pm$0.091\textsuperscript{7} & 0.747$\pm$0.153\textsuperscript{3} & 0.675$\pm$0.165\textsuperscript{5} & 0.617$\pm$0.168\textsuperscript{8} & 0.757$\pm$0.128\textsuperscript{2} & 0.747$\pm$0.153\textsuperscript{3} & \textbf{0.787$\pm$0.129\textsuperscript{1}} \\
0.244$\pm$0.232\textsuperscript{7} & 0.147$\pm$0.134\textsuperscript{8} & 0.316$\pm$0.289\textsuperscript{5} & 0.301$\pm$0.278\textsuperscript{6} & \textbf{0.537$\pm$0.491\textsuperscript{1}} & 0.382$\pm$0.351\textsuperscript{4} & 0.392$\pm$0.368\textsuperscript{3} & 0.513$\pm$0.470\textsuperscript{2} \\
0.560$\pm$0.009\textsuperscript{7} & 0.517$\pm$0.002\textsuperscript{8} & 0.649$\pm$0.032\textsuperscript{6} & 0.846$\pm$0.197\textsuperscript{3} & 0.905$\pm$0.023\textsuperscript{2} & 0.667$\pm$0.036\textsuperscript{4} & 0.659$\pm$0.042\textsuperscript{5} & \textbf{0.986$\pm$0.008\textsuperscript{1}} \\
\bottomrule
\end{tabular}}
\vspace{1em}

\caption{Ablation analysis results of the proposed model on Macro-F1, G-Mean, and AUPRC.}
\label{tab:ablationprog}
\resizebox{\textwidth}{!}{%
\begin{tabular}{@{}cccccccccccccccc@{}}
\toprule
 & \multicolumn{5}{c}{Macro-F1} & \multicolumn{5}{c}{G-Mean} & \multicolumn{5}{c}{AUPRC} \\
\cmidrule(lr){2-6}\cmidrule(lr){7-11}\cmidrule(lr){12-16}
Dataset & Base & DF & CF & DF+CF & Full & Base & DF & CF & DF+CF & Full & Base & DF & CF & DF+CF & Full \\
\midrule
abalone-17\_vs\_7-8-9-10 & 0.622$\pm$0.031 & 0.656$\pm$0.056 & 0.636$\pm$0.040 & 0.632$\pm$0.035 & \textbf{0.669$\pm$0.026} & 0.501$\pm$0.063 & 0.504$\pm$0.070 & 0.452$\pm$0.054 & 0.484$\pm$0.076 & \textbf{0.670$\pm$0.046} & 0.619$\pm$0.051 & 0.646$\pm$0.049 & 0.639$\pm$0.025 & 0.655$\pm$0.051 & \textbf{0.666$\pm$0.018} \\
abalone-20\_vs\_8-9-10 & 0.552$\pm$0.094 & 0.586$\pm$0.125 & 0.589$\pm$0.101 & 0.589$\pm$0.101 & \textbf{0.660$\pm$0.109} & 0.242$\pm$0.351 & 0.280$\pm$0.387 & 0.357$\pm$0.346 & 0.357$\pm$0.346 & \textbf{0.561$\pm$0.352} & 0.557$\pm$0.066 & 0.636$\pm$0.131 & 0.564$\pm$0.063 & 0.573$\pm$0.061 & \textbf{0.708$\pm$0.119} \\
abalone-21\_vs\_8 & 0.733$\pm$0.150 & 0.721$\pm$0.150 & 0.733$\pm$0.150 & 0.733$\pm$0.150 & \textbf{0.853$\pm$0.106} & 0.638$\pm$0.386 & 0.591$\pm$0.373 & 0.638$\pm$0.386 & 0.638$\pm$0.386 & \textbf{0.884$\pm$0.096} & 0.691$\pm$0.073 & 0.733$\pm$0.116 & 0.691$\pm$0.073 & 0.691$\pm$0.073 & \textbf{0.822$\pm$0.116} \\
abalone9-18 & 0.690$\pm$0.141 & 0.698$\pm$0.196 & 0.690$\pm$0.141 & 0.690$\pm$0.141 & \textbf{0.732$\pm$0.135} & 0.603$\pm$0.213 & 0.546$\pm$0.366 & 0.603$\pm$0.213 & 0.603$\pm$0.213 & \textbf{0.737$\pm$0.188} & 0.649$\pm$0.168 & 0.717$\pm$0.186 & 0.649$\pm$0.168 & 0.649$\pm$0.168 & \textbf{0.787$\pm$0.129} \\
autos & 0.799$\pm$0.099 & 0.815$\pm$0.042 & 0.814$\pm$0.092 & 0.804$\pm$0.105 & \textbf{0.841$\pm$0.021} & 0.666$\pm$0.377 & 0.781$\pm$0.040 & 0.671$\pm$0.380 & 0.671$\pm$0.380 & \textbf{0.826$\pm$0.050} & 0.478$\pm$0.437 & \textbf{0.537$\pm$0.490} & 0.477$\pm$0.436 & 0.473$\pm$0.433 & 0.513$\pm$0.470 \\
\bottomrule
\end{tabular}}
\end{sidewaystable}

\begin{sidewaystable}[p]
\caption{Sensitivity of the proposed model to the mutual-neighbor count $k$ on Macro-F1, G-Mean, and AUPRC.}
\label{tab:senskabs_k}
\resizebox{\textwidth}{!}{%
\begin{tabular}{@{}cccccccccccccccc@{}}
\toprule
 & \multicolumn{5}{c}{Macro-F1} & \multicolumn{5}{c}{G-Mean} & \multicolumn{5}{c}{AUPRC} \\
\cmidrule(lr){2-6}\cmidrule(lr){7-11}\cmidrule(lr){12-16}
Dataset & 2 & 5 & 10 & 15 & 20 & 2 & 5 & 10 & 15 & 20 & 2 & 5 & 10 & 15 & 20 \\
\midrule
abalone-17\_vs\_7-8-9-10 & 0.675$\pm$0.036 & \textbf{0.684$\pm$0.023} & 0.664$\pm$0.022 & 0.655$\pm$0.021 & 0.667$\pm$0.023 & 0.656$\pm$0.068 & \textbf{0.660$\pm$0.051} & 0.645$\pm$0.057 & 0.632$\pm$0.049 & 0.659$\pm$0.059 & 0.667$\pm$0.034 & 0.676$\pm$0.023 & 0.663$\pm$0.041 & 0.659$\pm$0.054 & 0.663$\pm$0.021 \\
abalone-20\_vs\_8-9-10 & 0.649$\pm$0.110 & 0.619$\pm$0.082 & 0.655$\pm$0.096 & 0.647$\pm$0.102 & 0.656$\pm$0.107 & 0.484$\pm$0.295 & 0.483$\pm$0.294 & 0.521$\pm$0.300 & 0.535$\pm$0.344 & 0.538$\pm$0.329 & 0.643$\pm$0.091 & 0.666$\pm$0.099 & 0.663$\pm$0.099 & 0.678$\pm$0.106 & 0.677$\pm$0.116 \\
abalone-21\_vs\_8 & 0.781$\pm$0.101 & 0.813$\pm$0.052 & 0.808$\pm$0.067 & \textbf{0.853$\pm$0.106} & 0.787$\pm$0.093 & 0.743$\pm$0.107 & 0.847$\pm$0.078 & 0.846$\pm$0.076 & \textbf{0.884$\pm$0.096} & 0.798$\pm$0.143 & 0.772$\pm$0.063 & 0.794$\pm$0.037 & 0.791$\pm$0.060 & \textbf{0.822$\pm$0.116} & 0.781$\pm$0.056 \\
abalone9-18 & 0.723$\pm$0.084 & 0.673$\pm$0.164 & 0.732$\pm$0.135 & \textbf{0.735$\pm$0.118} & 0.690$\pm$0.142 & 0.772$\pm$0.111 & 0.639$\pm$0.240 & 0.737$\pm$0.188 & 0.717$\pm$0.141 & 0.634$\pm$0.211 & 0.745$\pm$0.131 & 0.706$\pm$0.171 & \textbf{0.787$\pm$0.129} & 0.768$\pm$0.158 & 0.745$\pm$0.144 \\
autos & 0.815$\pm$0.072 & \textbf{0.881$\pm$0.042} & 0.842$\pm$0.052 & 0.835$\pm$0.022 & 0.853$\pm$0.023 & 0.795$\pm$0.097 & \textbf{0.870$\pm$0.048} & 0.827$\pm$0.079 & 0.818$\pm$0.061 & 0.845$\pm$0.020 & 0.511$\pm$0.468 & 0.511$\pm$0.469 & 0.513$\pm$0.471 & 0.519$\pm$0.475 & 0.506$\pm$0.464 \\
\bottomrule
\end{tabular}}
\vspace{1em}
\caption{Sensitivity of the proposed model to the density threshold $\rho_0$ on Macro-F1, G-Mean, and AUPRC.}
\label{tab:senskabs_rho}
\resizebox{\textwidth}{!}{%
\begin{tabular}{@{}ccccccccccccccccccc@{}}
\toprule
 & \multicolumn{6}{c}{Macro-F1} & \multicolumn{6}{c}{G-Mean} & \multicolumn{6}{c}{AUPRC} \\
\cmidrule(lr){2-7}\cmidrule(lr){8-13}\cmidrule(lr){14-19}
Dataset & 0.1 & 0.2 & 0.3 & 0.5 & 0.7 & 0.9 & 0.1 & 0.2 & 0.3 & 0.5 & 0.7 & 0.9 & 0.1 & 0.2 & 0.3 & 0.5 & 0.7 & 0.9 \\
\midrule
abalone-17\_vs\_7-8-9-10 & 0.659$\pm$0.034 & 0.678$\pm$0.047 & 0.656$\pm$0.031 & 0.664$\pm$0.019 & 0.658$\pm$0.022 & 0.659$\pm$0.040 & 0.536$\pm$0.075 & 0.621$\pm$0.051 & 0.582$\pm$0.014 & 0.659$\pm$0.060 & 0.632$\pm$0.084 & 0.617$\pm$0.096 & 0.688$\pm$0.024 & \textbf{0.689$\pm$0.052} & 0.678$\pm$0.020 & 0.683$\pm$0.033 & 0.637$\pm$0.028 & 0.661$\pm$0.015 \\
abalone-20\_vs\_8-9-10 & 0.609$\pm$0.108 & 0.673$\pm$0.169 & \textbf{0.689$\pm$0.122} & 0.626$\pm$0.084 & 0.660$\pm$0.109 & 0.671$\pm$0.129 & 0.367$\pm$0.336 & 0.474$\pm$0.438 & \textbf{0.573$\pm$0.343} & 0.456$\pm$0.265 & 0.561$\pm$0.352 & 0.536$\pm$0.346 & 0.617$\pm$0.096 & 0.673$\pm$0.133 & 0.698$\pm$0.110 & 0.668$\pm$0.105 & \textbf{0.708$\pm$0.119} & 0.695$\pm$0.116 \\
abalone-21\_vs\_8 & 0.656$\pm$0.160 & 0.791$\pm$0.038 & 0.780$\pm$0.058 & 0.839$\pm$0.083 & 0.811$\pm$0.096 & 0.782$\pm$0.084 & 0.475$\pm$0.458 & 0.845$\pm$0.077 & 0.844$\pm$0.073 & 0.883$\pm$0.095 & 0.881$\pm$0.097 & 0.798$\pm$0.144 & 0.695$\pm$0.103 & 0.778$\pm$0.059 & 0.758$\pm$0.075 & 0.811$\pm$0.117 & 0.797$\pm$0.091 & 0.794$\pm$0.037 \\
abalone9-18 & 0.720$\pm$0.143 & 0.722$\pm$0.149 & 0.732$\pm$0.135 & 0.732$\pm$0.107 & 0.679$\pm$0.155 & 0.717$\pm$0.101 & 0.668$\pm$0.218 & 0.685$\pm$0.254 & 0.737$\pm$0.188 & \textbf{0.774$\pm$0.105} & 0.668$\pm$0.234 & 0.743$\pm$0.133 & 0.725$\pm$0.158 & 0.767$\pm$0.186 & \textbf{0.787$\pm$0.129} & 0.757$\pm$0.124 & 0.767$\pm$0.149 & 0.746$\pm$0.110 \\
\bottomrule
\end{tabular}}
\end{sidewaystable}

As shown in Table \ref{tab:senskabs_k}, the best performance appears at $k=10$ on all three metrics, with Macro-F1, G-Mean, and AUPRC values of 0.791, 0.717, and 0.789, respectively. The results at $k=5$ and $k=15$ remain close, and performance drops only when $k$ becomes too small or too large. As shown in Table \ref{tab:senskabs_rho}, performance remains stable for $\rho_0\in[0.3, 0.9]$, with the three metrics around 0.786--0.797, 0.702--0.721, and 0.775--0.789, respectively, but clearly degrades at $\rho_0=0.1$.

These observations agree with the model mechanism. A too-small $k$ makes density estimates sensitive to local noise, whereas a too-large $k$ oversmooths the neighborhood structure; thus, a moderate $k$ is preferred. A too-small $\rho_0$ labels most minority samples as low-density samples and assigns them to boundary or noise regions. Region-guided generation then operates in low-confidence regions and may filter out too many samples, degrading the quality of the generated data.

\subsection{Collaborative Optimization Validation Experiment}

Although Section \ref{sec:Ablation Analysis} reports ablation experiments, the model couples weight updates with dynamic sampling, so the observed improvement could in principle come from the resampling module alone rather than from the coupling mechanism. To address this issue, we conduct five controlled comparisons, each removing one link from the closed loop. Tables \ref{tab:coord_macrof1}, \ref{tab:coord_gmean}, and \ref{tab:coord_auprc} report the five variants on Macro-F1, G-Mean, and AUPRC, respectively.

We define the five variants as follows. "w/o adapt" keeps the sampling strategy but reverts the weight update to standard AdaBoost. "w/o conf" and "w/o dens" remove the corresponding factor from both the weight update and region partitioning. "w/o guided" replaces boundary-to-dense guided interpolation with random intra-cluster interpolation. "w/o sampl" replaces the progressive scheduler with static resampling of a fixed quantity per epoch. Each row reports the five-fold mean and standard deviation for one dataset. The last row, "Avg (gap)", reports the average value of each variant over the 40 datasets, with the parenthesized value indicating the gap relative to the full method.

\begin{table}[H]
\caption{Macro-F1 results of the collaborative optimization validation experiment.}
\label{tab:coord_macrof1}
\small
\scriptsize\renewcommand{\arraystretch}{0.6}
\resizebox{\textwidth}{!}{%
\begin{tabular}{@{}ccccccc@{}}
\toprule
Dataset & Ours & w/o adapt & w/o conf & w/o dens & w/o guided & w/o sampl \\
\midrule
abalone-17\_vs\_7-8-9-10 & 0.669$\pm$0.026 & 0.681$\pm$0.043 & 0.655$\pm$0.030 & 0.653$\pm$0.035 & 0.671$\pm$0.039 & 0.632$\pm$0.035 \\
abalone-20\_vs\_8-9-10 & 0.660$\pm$0.109 & 0.620$\pm$0.120 & 0.632$\pm$0.087 & 0.640$\pm$0.141 & 0.658$\pm$0.114 & 0.589$\pm$0.101 \\
abalone-21\_vs\_8 & 0.853$\pm$0.106 & 0.789$\pm$0.113 & 0.754$\pm$0.102 & 0.800$\pm$0.050 & 0.756$\pm$0.099 & 0.733$\pm$0.150 \\
autos & 0.841$\pm$0.021 & 0.834$\pm$0.031 & 0.799$\pm$0.061 & 0.840$\pm$0.028 & 0.830$\pm$0.034 & 0.804$\pm$0.105 \\
car-good & 0.957$\pm$0.009 & 0.894$\pm$0.036 & 0.895$\pm$0.032 & 0.938$\pm$0.033 & 0.911$\pm$0.031 & 0.957$\pm$0.009 \\
car-vgood & 0.991$\pm$0.019 & 1.000$\pm$0.000 & 0.962$\pm$0.040 & 1.000$\pm$0.000 & 0.996$\pm$0.009 & 0.978$\pm$0.031 \\
cleveland-0\_vs\_4 & 0.743$\pm$0.069 & 0.765$\pm$0.093 & 0.728$\pm$0.197 & 0.788$\pm$0.128 & 0.773$\pm$0.145 & 0.677$\pm$0.052 \\
dermatology & 0.925$\pm$0.020 & 0.915$\pm$0.022 & 0.959$\pm$0.021 & 0.917$\pm$0.026 & 0.918$\pm$0.026 & 0.926$\pm$0.023 \\
ecoli-0-1-4-7\_vs\_5-6 & 0.863$\pm$0.047 & 0.851$\pm$0.032 & 0.857$\pm$0.083 & 0.831$\pm$0.047 & 0.842$\pm$0.073 & 0.826$\pm$0.043 \\
ecoli-0-2-6-7\_vs\_3-5 & 0.819$\pm$0.131 & 0.823$\pm$0.083 & 0.835$\pm$0.106 & 0.835$\pm$0.106 & 0.870$\pm$0.106 & 0.819$\pm$0.131 \\
ecoli-0-3-4-7\_vs\_5-6 & 0.891$\pm$0.039 & 0.871$\pm$0.063 & 0.798$\pm$0.046 & 0.884$\pm$0.051 & 0.854$\pm$0.118 & 0.817$\pm$0.039 \\
ecoli-0-6-7\_vs\_3-5 & 0.837$\pm$0.129 & 0.795$\pm$0.106 & 0.809$\pm$0.095 & 0.857$\pm$0.137 & 0.846$\pm$0.132 & 0.861$\pm$0.154 \\
ecoli-0-6-7\_vs\_5 & 0.914$\pm$0.057 & 0.904$\pm$0.070 & 0.814$\pm$0.096 & 0.866$\pm$0.117 & 0.914$\pm$0.057 & 0.847$\pm$0.125 \\
glass & 0.706$\pm$0.052 & 0.672$\pm$0.049 & 0.708$\pm$0.097 & 0.656$\pm$0.115 & 0.688$\pm$0.068 & 0.694$\pm$0.089 \\
glass-0-1-4-6\_vs\_2 & 0.705$\pm$0.149 & 0.684$\pm$0.130 & 0.683$\pm$0.145 & 0.693$\pm$0.139 & 0.694$\pm$0.156 & 0.609$\pm$0.152 \\
glass-0-1-6\_vs\_2 & 0.635$\pm$0.129 & 0.593$\pm$0.085 & 0.520$\pm$0.085 & 0.553$\pm$0.102 & 0.556$\pm$0.090 & 0.535$\pm$0.118 \\
glass-0-6\_vs\_5 & 0.928$\pm$0.098 & 0.928$\pm$0.098 & 0.873$\pm$0.122 & 0.928$\pm$0.098 & 0.964$\pm$0.080 & 0.928$\pm$0.098 \\
glass1 & 0.733$\pm$0.064 & 0.730$\pm$0.047 & 0.698$\pm$0.044 & 0.751$\pm$0.076 & 0.710$\pm$0.096 & 0.745$\pm$0.052 \\
glass4 & 0.892$\pm$0.147 & 0.858$\pm$0.136 & 0.858$\pm$0.136 & 0.835$\pm$0.169 & 0.699$\pm$0.155 & 0.827$\pm$0.153 \\
glass6 & 0.919$\pm$0.030 & 0.921$\pm$0.069 & 0.951$\pm$0.048 & 0.703$\pm$0.244 & 0.919$\pm$0.030 & 0.919$\pm$0.030 \\
kddcup-land\_vs\_satan & 1.000$\pm$0.000 & 1.000$\pm$0.000 & 0.989$\pm$0.025 & 1.000$\pm$0.000 & 0.989$\pm$0.025 & 1.000$\pm$0.000 \\
kr-vs-k-three\_vs\_eleven & 0.997$\pm$0.007 & 0.997$\pm$0.007 & 1.000$\pm$0.000 & 0.997$\pm$0.007 & 0.997$\pm$0.007 & 0.997$\pm$0.007 \\
kr-vs-k-zero-one\_vs\_draw & 0.985$\pm$0.011 & 0.975$\pm$0.019 & 0.972$\pm$0.016 & 0.985$\pm$0.011 & 0.975$\pm$0.019 & 0.978$\pm$0.020 \\
kr-vs-k-zero\_vs\_eight & 1.000$\pm$0.000 & 1.000$\pm$0.000 & 1.000$\pm$0.000 & 1.000$\pm$0.000 & 0.992$\pm$0.018 & 1.000$\pm$0.000 \\
pageblocks & 0.736$\pm$0.082 & 0.720$\pm$0.087 & 0.825$\pm$0.091 & 0.710$\pm$0.167 & 0.710$\pm$0.080 & 0.766$\pm$0.129 \\
poker-8-9\_vs\_6 & 0.966$\pm$0.076 & 0.966$\pm$0.076 & 0.927$\pm$0.054 & 0.955$\pm$0.074 & 0.975$\pm$0.057 & 0.622$\pm$0.130 \\
poker-8\_vs\_6 & 0.749$\pm$0.251 & 0.779$\pm$0.260 & 0.799$\pm$0.201 & 0.756$\pm$0.252 & 0.738$\pm$0.252 & 0.597$\pm$0.225 \\
shuttle & 0.959$\pm$0.090 & 0.908$\pm$0.126 & 0.959$\pm$0.090 & 0.959$\pm$0.090 & 0.925$\pm$0.082 & 0.957$\pm$0.089 \\
vehicle1 & 0.733$\pm$0.059 & 0.695$\pm$0.057 & 0.717$\pm$0.056 & 0.722$\pm$0.017 & 0.719$\pm$0.049 & 0.693$\pm$0.060 \\
vowel0 & 0.966$\pm$0.022 & 0.955$\pm$0.031 & 0.948$\pm$0.037 & 0.955$\pm$0.029 & 0.943$\pm$0.038 & 0.955$\pm$0.027 \\
winequality-red-4 & 0.561$\pm$0.036 & 0.583$\pm$0.077 & 0.584$\pm$0.045 & 0.542$\pm$0.057 & 0.579$\pm$0.034 & 0.550$\pm$0.048 \\
winequality-red-8\_vs\_6 & 0.677$\pm$0.045 & 0.618$\pm$0.060 & 0.670$\pm$0.117 & 0.667$\pm$0.076 & 0.582$\pm$0.133 & 0.582$\pm$0.087 \\
winequality-red-8\_vs\_6-7 & 0.566$\pm$0.085 & 0.545$\pm$0.077 & 0.563$\pm$0.106 & 0.579$\pm$0.087 & 0.661$\pm$0.138 & 0.544$\pm$0.082 \\
yeast & 0.489$\pm$0.055 & 0.500$\pm$0.018 & 0.478$\pm$0.053 & 0.492$\pm$0.025 & 0.486$\pm$0.029 & 0.397$\pm$0.039 \\
yeast-0-2-5-7-9\_vs\_3-6-8 & 0.882$\pm$0.024 & 0.883$\pm$0.029 & 0.886$\pm$0.016 & 0.895$\pm$0.022 & 0.885$\pm$0.022 & 0.872$\pm$0.026 \\
yeast-1\_vs\_7 & 0.699$\pm$0.092 & 0.741$\pm$0.073 & 0.621$\pm$0.089 & 0.677$\pm$0.102 & 0.591$\pm$0.087 & 0.647$\pm$0.034 \\
yeast-2\_vs\_4 & 0.850$\pm$0.050 & 0.854$\pm$0.032 & 0.847$\pm$0.064 & 0.839$\pm$0.048 & 0.864$\pm$0.042 & 0.835$\pm$0.071 \\
yeast-2\_vs\_8 & 0.821$\pm$0.101 & 0.809$\pm$0.068 & 0.846$\pm$0.055 & 0.741$\pm$0.156 & 0.829$\pm$0.086 & 0.829$\pm$0.086 \\
yeast1 & 0.716$\pm$0.022 & 0.689$\pm$0.018 & 0.701$\pm$0.022 & 0.710$\pm$0.029 & 0.703$\pm$0.026 & 0.693$\pm$0.021 \\
zoo-3 & 0.654$\pm$0.243 & 0.582$\pm$0.234 & 0.723$\pm$0.227 & 0.582$\pm$0.234 & 0.579$\pm$0.132 & 0.648$\pm$0.248 \\
\midrule
Avg (gap) & 0.8122 & 0.7982 $(-0.014)$ & 0.7961 $(-0.016)$ & 0.7933 $(-0.019)$ & 0.7947 $(-0.017)$ & 0.7721 $(-0.040)$ \\
\bottomrule
\end{tabular}}
\end{table}

\begin{table}[H]
\caption{G-Mean results of the collaborative optimization validation experiment.}
\label{tab:coord_gmean}
\small
\scriptsize\renewcommand{\arraystretch}{0.6}
\resizebox{\textwidth}{!}{%
\begin{tabular}{@{}ccccccc@{}}
\toprule
Dataset & Ours & w/o adapt & w/o conf & w/o dens & w/o guided & w/o sampl \\
\midrule
abalone-17\_vs\_7-8-9-10 & 0.670$\pm$0.046 & 0.657$\pm$0.101 & 0.667$\pm$0.058 & 0.626$\pm$0.098 & 0.671$\pm$0.048 & 0.484$\pm$0.076 \\
abalone-20\_vs\_8-9-10 & 0.561$\pm$0.352 & 0.418$\pm$0.399 & 0.519$\pm$0.299 & 0.447$\pm$0.423 & 0.548$\pm$0.319 & 0.357$\pm$0.346 \\
abalone-21\_vs\_8 & 0.884$\pm$0.096 & 0.788$\pm$0.053 & 0.786$\pm$0.053 & 0.881$\pm$0.097 & 0.795$\pm$0.142 & 0.638$\pm$0.386 \\
autos & 0.826$\pm$0.050 & 0.814$\pm$0.069 & 0.781$\pm$0.054 & 0.828$\pm$0.057 & 0.808$\pm$0.069 & 0.671$\pm$0.380 \\
car-good & 0.939$\pm$0.021 & 0.865$\pm$0.038 & 0.904$\pm$0.022 & 0.922$\pm$0.039 & 0.882$\pm$0.049 & 0.939$\pm$0.021 \\
car-vgood & 0.984$\pm$0.036 & 1.000$\pm$0.000 & 0.943$\pm$0.048 & 1.000$\pm$0.000 & 0.992$\pm$0.018 & 0.959$\pm$0.058 \\
cleveland-0\_vs\_4 & 0.741$\pm$0.145 & 0.766$\pm$0.155 & 0.611$\pm$0.377 & 0.803$\pm$0.186 & 0.748$\pm$0.162 & 0.655$\pm$0.096 \\
dermatology & 0.912$\pm$0.042 & 0.903$\pm$0.045 & 0.957$\pm$0.018 & 0.900$\pm$0.046 & 0.903$\pm$0.046 & 0.916$\pm$0.034 \\
ecoli-0-1-4-7\_vs\_5-6 & 0.878$\pm$0.069 & 0.877$\pm$0.073 & 0.856$\pm$0.096 & 0.849$\pm$0.077 & 0.832$\pm$0.100 & 0.808$\pm$0.093 \\
ecoli-0-2-6-7\_vs\_3-5 & 0.738$\pm$0.196 & 0.756$\pm$0.113 & 0.777$\pm$0.149 & 0.777$\pm$0.149 & 0.829$\pm$0.131 & 0.738$\pm$0.196 \\
ecoli-0-3-4-7\_vs\_5-6 & 0.900$\pm$0.088 & 0.878$\pm$0.082 & 0.758$\pm$0.089 & 0.898$\pm$0.088 & 0.818$\pm$0.211 & 0.806$\pm$0.055 \\
ecoli-0-6-7\_vs\_3-5 & 0.784$\pm$0.206 & 0.751$\pm$0.175 & 0.755$\pm$0.177 & 0.835$\pm$0.237 & 0.832$\pm$0.235 & 0.836$\pm$0.242 \\
ecoli-0-6-7\_vs\_5 & 0.888$\pm$0.063 & 0.886$\pm$0.064 & 0.804$\pm$0.181 & 0.815$\pm$0.187 & 0.888$\pm$0.063 & 0.784$\pm$0.191 \\
glass & 0.687$\pm$0.087 & 0.486$\pm$0.278 & 0.585$\pm$0.338 & 0.450$\pm$0.411 & 0.568$\pm$0.320 & 0.547$\pm$0.316 \\
glass-0-1-4-6\_vs\_2 & 0.624$\pm$0.353 & 0.576$\pm$0.338 & 0.575$\pm$0.340 & 0.595$\pm$0.336 & 0.580$\pm$0.343 & 0.394$\pm$0.377 \\
glass-0-1-6\_vs\_2 & 0.462$\pm$0.270 & 0.450$\pm$0.259 & 0.207$\pm$0.284 & 0.306$\pm$0.280 & 0.306$\pm$0.281 & 0.252$\pm$0.359 \\
glass-0-6\_vs\_5 & 0.936$\pm$0.129 & 0.936$\pm$0.129 & 0.874$\pm$0.161 & 0.936$\pm$0.129 & 0.995$\pm$0.011 & 0.936$\pm$0.129 \\
glass1 & 0.723$\pm$0.059 & 0.718$\pm$0.057 & 0.681$\pm$0.036 & 0.742$\pm$0.056 & 0.688$\pm$0.096 & 0.730$\pm$0.038 \\
glass4 & 0.873$\pm$0.189 & 0.814$\pm$0.185 & 0.814$\pm$0.185 & 0.768$\pm$0.220 & 0.582$\pm$0.367 & 0.822$\pm$0.233 \\
glass6 & 0.916$\pm$0.041 & 0.883$\pm$0.107 & 0.943$\pm$0.053 & 0.508$\pm$0.494 & 0.916$\pm$0.041 & 0.916$\pm$0.041 \\
kddcup-land\_vs\_satan & 1.000$\pm$0.000 & 1.000$\pm$0.000 & 0.979$\pm$0.047 & 1.000$\pm$0.000 & 0.979$\pm$0.047 & 1.000$\pm$0.000 \\
kr-vs-k-three\_vs\_eleven & 0.994$\pm$0.014 & 0.994$\pm$0.014 & 1.000$\pm$0.000 & 0.994$\pm$0.014 & 0.994$\pm$0.014 & 0.994$\pm$0.014 \\
kr-vs-k-zero-one\_vs\_draw & 0.976$\pm$0.017 & 0.975$\pm$0.018 & 0.965$\pm$0.028 & 0.976$\pm$0.017 & 0.970$\pm$0.027 & 0.975$\pm$0.030 \\
kr-vs-k-zero\_vs\_eight & 1.000$\pm$0.000 & 1.000$\pm$0.000 & 1.000$\pm$0.000 & 1.000$\pm$0.000 & 1.000$\pm$0.001 & 1.000$\pm$0.000 \\
pageblocks & 0.502$\pm$0.459 & 0.482$\pm$0.441 & 0.676$\pm$0.385 & 0.644$\pm$0.364 & 0.450$\pm$0.414 & 0.678$\pm$0.380 \\
poker-8-9\_vs\_6 & 0.955$\pm$0.101 & 0.955$\pm$0.101 & 0.868$\pm$0.095 & 0.934$\pm$0.101 & 0.955$\pm$0.101 & 0.342$\pm$0.321 \\
poker-8\_vs\_6 & 0.515$\pm$0.501 & 0.563$\pm$0.520 & 0.627$\pm$0.394 & 0.563$\pm$0.519 & 0.515$\pm$0.501 & 0.200$\pm$0.447 \\
shuttle & 0.799$\pm$0.447 & 0.599$\pm$0.547 & 0.799$\pm$0.447 & 0.799$\pm$0.447 & 0.767$\pm$0.434 & 0.797$\pm$0.446 \\
vehicle1 & 0.724$\pm$0.076 & 0.674$\pm$0.075 & 0.706$\pm$0.074 & 0.701$\pm$0.043 & 0.704$\pm$0.054 & 0.653$\pm$0.073 \\
vowel0 & 0.972$\pm$0.051 & 0.970$\pm$0.052 & 0.954$\pm$0.057 & 0.970$\pm$0.051 & 0.936$\pm$0.071 & 0.939$\pm$0.055 \\
winequality-red-4 & 0.409$\pm$0.113 & 0.462$\pm$0.147 & 0.422$\pm$0.075 & 0.322$\pm$0.215 & 0.458$\pm$0.093 & 0.295$\pm$0.177 \\
winequality-red-8\_vs\_6 & 0.637$\pm$0.138 & 0.591$\pm$0.147 & 0.526$\pm$0.324 & 0.606$\pm$0.092 & 0.280$\pm$0.383 & 0.313$\pm$0.287 \\
winequality-red-8\_vs\_6-7 & 0.336$\pm$0.318 & 0.238$\pm$0.333 & 0.240$\pm$0.336 & 0.313$\pm$0.287 & 0.540$\pm$0.324 & 0.198$\pm$0.272 \\
yeast & 0.100$\pm$0.225 & 0.091$\pm$0.204 & 0.000$\pm$0.000 & 0.093$\pm$0.207 & 0.000$\pm$0.000 & 0.000$\pm$0.000 \\
yeast-0-2-5-7-9\_vs\_3-6-8 & 0.866$\pm$0.054 & 0.877$\pm$0.030 & 0.873$\pm$0.030 & 0.875$\pm$0.031 & 0.868$\pm$0.044 & 0.854$\pm$0.027 \\
yeast-1\_vs\_7 & 0.680$\pm$0.123 & 0.725$\pm$0.148 & 0.514$\pm$0.161 & 0.653$\pm$0.134 & 0.458$\pm$0.290 & 0.627$\pm$0.099 \\
yeast-2\_vs\_4 & 0.873$\pm$0.078 & 0.849$\pm$0.048 & 0.835$\pm$0.079 & 0.841$\pm$0.089 & 0.903$\pm$0.088 & 0.842$\pm$0.090 \\
yeast-2\_vs\_8 & 0.728$\pm$0.151 & 0.697$\pm$0.129 & 0.769$\pm$0.086 & 0.555$\pm$0.336 & 0.728$\pm$0.150 & 0.728$\pm$0.150 \\
yeast1 & 0.692$\pm$0.025 & 0.664$\pm$0.024 & 0.688$\pm$0.025 & 0.686$\pm$0.034 & 0.684$\pm$0.038 & 0.645$\pm$0.020 \\
zoo-3 & 0.395$\pm$0.541 & 0.200$\pm$0.447 & 0.590$\pm$0.538 & 0.200$\pm$0.447 & 0.379$\pm$0.519 & 0.395$\pm$0.541 \\
\midrule
Avg (gap) & 0.7520 & 0.7207 $(-0.031)$ & 0.7207 $(-0.031)$ & 0.7152 $(-0.037)$ & 0.7187 $(-0.033)$ & 0.6666 $(-0.085)$ \\
\bottomrule
\end{tabular}}
\end{table}

\begin{table}[H]
\caption{AUPRC results of the collaborative optimization validation experiment.}
\label{tab:coord_auprc}

\scriptsize\renewcommand{\arraystretch}{0.6}
\resizebox{\textwidth}{!}{%
\begin{tabular}{@{}ccccccc@{}}
\toprule
Dataset & Ours & w/o adapt & w/o conf & w/o dens & w/o guided & w/o sampl \\
\midrule
abalone-17\_vs\_7-8-9-10 & 0.666$\pm$0.018 & 0.654$\pm$0.026 & 0.645$\pm$0.011 & 0.647$\pm$0.024 & 0.671$\pm$0.025 & 0.655$\pm$0.051 \\
abalone-20\_vs\_8-9-10 & 0.708$\pm$0.119 & 0.671$\pm$0.113 & 0.673$\pm$0.112 & 0.686$\pm$0.124 & 0.667$\pm$0.108 & 0.573$\pm$0.061 \\
abalone-21\_vs\_8 & 0.822$\pm$0.116 & 0.789$\pm$0.114 & 0.814$\pm$0.092 & 0.797$\pm$0.044 & 0.775$\pm$0.059 & 0.691$\pm$0.073 \\
autos & 0.513$\pm$0.470 & 0.497$\pm$0.457 & 0.540$\pm$0.495 & 0.502$\pm$0.459 & 0.527$\pm$0.482 & 0.473$\pm$0.433 \\
car-good & 0.986$\pm$0.008 & 0.970$\pm$0.024 & 0.948$\pm$0.020 & 0.978$\pm$0.010 & 0.952$\pm$0.027 & 0.986$\pm$0.008 \\
car-vgood & 1.000$\pm$0.000 & 1.000$\pm$0.000 & 0.992$\pm$0.015 & 1.000$\pm$0.000 & 1.000$\pm$0.000 & 0.961$\pm$0.054 \\
cleveland-0\_vs\_4 & 0.809$\pm$0.129 & 0.738$\pm$0.084 & 0.745$\pm$0.161 & 0.801$\pm$0.156 & 0.755$\pm$0.161 & 0.607$\pm$0.069 \\
dermatology & 0.945$\pm$0.029 & 0.930$\pm$0.033 & 0.979$\pm$0.010 & 0.932$\pm$0.027 & 0.960$\pm$0.027 & 0.966$\pm$0.027 \\
ecoli-0-1-4-7\_vs\_5-6 & 0.887$\pm$0.093 & 0.897$\pm$0.074 & 0.869$\pm$0.072 & 0.886$\pm$0.061 & 0.843$\pm$0.100 & 0.744$\pm$0.064 \\
ecoli-0-2-6-7\_vs\_3-5 & 0.874$\pm$0.136 & 0.817$\pm$0.108 & 0.862$\pm$0.134 & 0.858$\pm$0.147 & 0.883$\pm$0.122 & 0.877$\pm$0.133 \\
ecoli-0-3-4-7\_vs\_5-6 & 0.893$\pm$0.090 & 0.856$\pm$0.077 & 0.855$\pm$0.073 & 0.894$\pm$0.091 & 0.896$\pm$0.096 & 0.753$\pm$0.057 \\
ecoli-0-6-7\_vs\_3-5 & 0.835$\pm$0.172 & 0.807$\pm$0.154 & 0.843$\pm$0.153 & 0.843$\pm$0.179 & 0.843$\pm$0.179 & 0.834$\pm$0.196 \\
ecoli-0-6-7\_vs\_5 & 0.880$\pm$0.079 & 0.880$\pm$0.079 & 0.882$\pm$0.097 & 0.880$\pm$0.079 & 0.880$\pm$0.079 & 0.784$\pm$0.158 \\
glass & 0.786$\pm$0.033 & 0.743$\pm$0.055 & 0.835$\pm$0.012 & 0.753$\pm$0.060 & 0.795$\pm$0.055 & 0.767$\pm$0.096 \\
glass-0-1-4-6\_vs\_2 & 0.706$\pm$0.127 & 0.717$\pm$0.139 & 0.735$\pm$0.144 & 0.703$\pm$0.132 & 0.714$\pm$0.125 & 0.591$\pm$0.140 \\
glass-0-1-6\_vs\_2 & 0.635$\pm$0.119 & 0.597$\pm$0.069 & 0.629$\pm$0.113 & 0.636$\pm$0.160 & 0.593$\pm$0.072 & 0.550$\pm$0.084 \\
glass-0-6\_vs\_5 & 0.950$\pm$0.112 & 0.950$\pm$0.112 & 0.933$\pm$0.109 & 0.950$\pm$0.112 & 0.950$\pm$0.112 & 0.950$\pm$0.112 \\
glass1 & 0.771$\pm$0.070 & 0.754$\pm$0.034 & 0.793$\pm$0.043 & 0.769$\pm$0.105 & 0.711$\pm$0.086 & 0.708$\pm$0.088 \\
glass4 & 0.880$\pm$0.166 & 0.872$\pm$0.176 & 0.855$\pm$0.164 & 0.863$\pm$0.155 & 0.855$\pm$0.200 & 0.766$\pm$0.174 \\
glass6 & 0.939$\pm$0.062 & 0.942$\pm$0.043 & 0.935$\pm$0.062 & 0.933$\pm$0.060 & 0.933$\pm$0.053 & 0.931$\pm$0.052 \\
kddcup-land\_vs\_satan & 1.000$\pm$0.000 & 1.000$\pm$0.000 & 1.000$\pm$0.000 & 1.000$\pm$0.000 & 1.000$\pm$0.000 & 1.000$\pm$0.000 \\
kr-vs-k-three\_vs\_eleven & 0.994$\pm$0.014 & 0.994$\pm$0.014 & 1.000$\pm$0.000 & 0.994$\pm$0.014 & 0.994$\pm$0.014 & 0.994$\pm$0.014 \\
kr-vs-k-zero-one\_vs\_draw & 0.985$\pm$0.013 & 0.976$\pm$0.027 & 0.990$\pm$0.013 & 0.985$\pm$0.013 & 0.968$\pm$0.036 & 0.972$\pm$0.026 \\
kr-vs-k-zero\_vs\_eight & 1.000$\pm$0.000 & 1.000$\pm$0.000 & 1.000$\pm$0.000 & 1.000$\pm$0.000 & 1.000$\pm$0.000 & 1.000$\pm$0.000 \\
pageblocks & 0.464$\pm$0.424 & 0.464$\pm$0.428 & 0.494$\pm$0.458 & 0.485$\pm$0.449 & 0.471$\pm$0.430 & 0.468$\pm$0.443 \\
poker-8-9\_vs\_6 & 0.974$\pm$0.058 & 0.961$\pm$0.088 & 0.959$\pm$0.073 & 0.960$\pm$0.090 & 0.961$\pm$0.088 & 0.782$\pm$0.191 \\
poker-8\_vs\_6 & 0.748$\pm$0.237 & 0.779$\pm$0.238 & 0.795$\pm$0.168 & 0.775$\pm$0.224 & 0.754$\pm$0.235 & 0.611$\pm$0.219 \\
shuttle & 0.359$\pm$0.497 & 0.359$\pm$0.497 & 0.399$\pm$0.547 & 0.359$\pm$0.497 & 0.359$\pm$0.497 & 0.359$\pm$0.497 \\
vehicle1 & 0.792$\pm$0.036 & 0.773$\pm$0.042 & 0.783$\pm$0.047 & 0.797$\pm$0.029 & 0.788$\pm$0.052 & 0.788$\pm$0.045 \\
vowel0 & 0.986$\pm$0.018 & 0.987$\pm$0.017 & 0.989$\pm$0.014 & 0.974$\pm$0.031 & 0.986$\pm$0.018 & 0.938$\pm$0.053 \\
winequality-red-4 & 0.556$\pm$0.013 & 0.560$\pm$0.029 & 0.574$\pm$0.037 & 0.548$\pm$0.027 & 0.561$\pm$0.019 & 0.525$\pm$0.030 \\
winequality-red-8\_vs\_6 & 0.685$\pm$0.062 & 0.637$\pm$0.090 & 0.694$\pm$0.110 & 0.675$\pm$0.109 & 0.636$\pm$0.131 & 0.664$\pm$0.099 \\
winequality-red-8\_vs\_6-7 & 0.612$\pm$0.126 & 0.594$\pm$0.114 & 0.594$\pm$0.115 & 0.597$\pm$0.076 & 0.673$\pm$0.157 & 0.557$\pm$0.062 \\
yeast & 0.546$\pm$0.041 & 0.534$\pm$0.033 & 0.539$\pm$0.047 & 0.526$\pm$0.028 & 0.527$\pm$0.032 & 0.329$\pm$0.049 \\
yeast-0-2-5-7-9\_vs\_3-6-8 & 0.905$\pm$0.030 & 0.918$\pm$0.040 & 0.903$\pm$0.024 & 0.900$\pm$0.044 & 0.919$\pm$0.023 & 0.910$\pm$0.031 \\
yeast-1\_vs\_7 & 0.714$\pm$0.100 & 0.696$\pm$0.101 & 0.670$\pm$0.088 & 0.695$\pm$0.097 & 0.655$\pm$0.087 & 0.582$\pm$0.030 \\
yeast-2\_vs\_4 & 0.914$\pm$0.066 & 0.876$\pm$0.050 & 0.908$\pm$0.056 & 0.905$\pm$0.060 & 0.902$\pm$0.045 & 0.842$\pm$0.083 \\
yeast-2\_vs\_8 & 0.767$\pm$0.080 & 0.772$\pm$0.088 & 0.792$\pm$0.054 & 0.757$\pm$0.094 & 0.786$\pm$0.096 & 0.774$\pm$0.090 \\
yeast1 & 0.769$\pm$0.019 & 0.760$\pm$0.023 & 0.747$\pm$0.024 & 0.758$\pm$0.015 & 0.750$\pm$0.019 & 0.746$\pm$0.024 \\
zoo-3 & 0.749$\pm$0.251 & 0.718$\pm$0.260 & 0.699$\pm$0.210 & 0.718$\pm$0.260 & 0.682$\pm$0.208 & 0.649$\pm$0.224 \\
\midrule
Avg (gap) & 0.8001 & 0.7860 $(-0.014)$ & 0.7973 $(-0.003)$ & 0.7930 $(-0.007)$ & 0.7893 $(-0.011)$ & 0.7414 $(-0.059)$ \\
\bottomrule
\end{tabular}}
\end{table}

The results show that static resampling causes by far the largest performance drop, reducing Macro-F1 by 0.040, G-Mean by 0.085, and AUPRC by 0.059. This degradation is roughly three times greater than that of any other variant and affects G-Mean most strongly, as G-Mean is especially sensitive to minority-class recall. Thus, the overall performance gain stems from dynamic, density-aware resampling within the loop rather than from a standalone oversampler. The remaining components also contribute consistently. The density factor primarily improves G-Mean by 0.037, while the confidence factor mainly improves Macro-F1 by 0.016 and G-Mean by 0.031, with AUPRC changing only slightly by 0.003. Adaptive weights and guided generation contribute more uniformly, with metric drops ranging from 0.014 to 0.033. Because every variant loses performance across all three metrics, no single component explains the improvement alone; instead, the performance gain originates from the closed-loop coupling.

\section{Conclusion}
\label{sec:Conclusion}

This study addresses a critical limitation in current research where imbalanced learning and model training are typically decoupled. We propose a novel collaborative optimization model: the Density-Aware and Region-Guided Boosting. Unlike traditional methods that treat data sampling and weight updates as isolated steps, we integrate these processes to establish a closed-loop synergy between imbalanced learning and ensemble construction.

Our primary contribution is the use of density factor and confidence factor as a bridge to link sample weight updates with dynamic sampling. This facilitates a unified training loop, the effectiveness of which is verified on multiple public imbalanced datasets. This research offers a promising perspective for future developments in the field. While we achieved strong performance using AdaBoost, this collaborative optimization strategy can be extended to other ensemble models. Such an extension would yield more flexible and adaptive unified models that are highly valuable for practical applications.

However, the proposed method still has limitations. Density estimation based on nearest neighbors is only an approximation and becomes computationally expensive as the dataset size increases. Furthermore, some models do not update sample weights based solely on classification error rates; for example, the GBDTs employ residual updates. Therefore, our future work will focus on two directions. First, we will investigate simpler and more accurate region partitioning mechanisms to reduce the impact of noise and class overlap. Second, we aim to extend our collaborative optimization strategy to the GBDTs and neural networks to enable broader practical applications.

\bibliographystyle{unsrtnat}
\nocite{*}
\bibliography{references} 

\begin{thebibliography}{35}
\providecommand{\natexlab}[1]{#1}
\providecommand{\url}[1]{\texttt{#1}}
\expandafter\ifx\csname urlstyle\endcsname\relax
  \providecommand{\doi}[1]{doi: #1}\else
  \providecommand{\doi}{doi: \begingroup \urlstyle{rm}\Url}\fi

\bibitem[Sharma and Verma(2026)]{1}
A.~K. Sharma and N.~K. Verma.
\newblock A novel vision transformer with selective residual in multihead self-attention for pattern recognition.
\newblock \emph{Pattern Recognition}, 172:\penalty0 112497, 2026.
\newblock \doi{10.1016/j.patcog.2025.112497}.

\bibitem[Han et~al.(2026)Han, Lei, and Li]{2}
X.~Han, M.~Lei, and J.~Li.
\newblock Hypergraph-based semantic and topological self-supervised learning for brain disease diagnosis.
\newblock \emph{Pattern Recognition}, 169:\penalty0 111921, 2026.
\newblock \doi{10.1016/j.patcog.2025.111921}.

\bibitem[Ding et~al.(2024)Ding, Zhong, Qin, Li, Fan, Deng, Ling, and Xiang]{3}
Z.~Ding, G.~Zhong, X.~Qin, Q.~Li, Z.~Fan, Z.~Deng, X.~Ling, and W.~Xiang.
\newblock {MF-Net}: Multi-frequency intrusion detection network for internet traffic data.
\newblock \emph{Pattern Recognition}, 146:\penalty0 109999, 2024.
\newblock \doi{10.1016/j.patcog.2023.109999}.

\bibitem[Yang et~al.(2026)Yang, Ge, Li, Li, and Wu]{4}
X.~Yang, J.~Ge, H.~Li, L.~Li, and B.~Wu.
\newblock {ReID}: Re-ranking through image description for object re-identification.
\newblock \emph{Pattern Recognition}, 172:\penalty0 112552, 2026.
\newblock \doi{10.1016/j.patcog.2025.112552}.

\bibitem[Zouhri et~al.(2024)Zouhri, Idri, and Ratnani]{5}
H.~Zouhri, A.~Idri, and A.~Ratnani.
\newblock Evaluating the impact of filter-based feature selection in intrusion detection systems.
\newblock \emph{International Journal of Information Security}, 23:\penalty0 759--785, 2024.
\newblock \doi{10.1007/s10207-023-00767-y}.

\bibitem[Ren et~al.(2025)Ren, Tang, Ren, et~al.]{6}
H.~Ren, Y.~Tang, S.~Ren, et~al.
\newblock {ADHS-EL}: Dynamic ensemble learning with adversarial augmentation for accurate and robust network intrusion detection.
\newblock \emph{Journal of King Saud University-Computer and Information Sciences}, 37:\penalty0 7, 2025.
\newblock \doi{10.1007/s44443-025-00015-4}.

\bibitem[Chawla et~al.(2002)Chawla, Bowyer, Hall, and Kegelmeyer]{7}
Nitesh~V Chawla, Kevin~W Bowyer, Lawrence~O Hall, and W~Philip Kegelmeyer.
\newblock {SMOTE}: Synthetic minority over-sampling technique.
\newblock \emph{Journal of Artificial Intelligence Research}, 16\penalty0 (1):\penalty0 321--357, 2002.

\bibitem[He and Garcia(2009)]{8}
Haibo He and Edwardo~A Garcia.
\newblock Learning from imbalanced data.
\newblock \emph{IEEE Transactions on Knowledge and Data Engineering}, 21\penalty0 (9):\penalty0 1263--1284, 2009.
\newblock \doi{10.1109/TKDE.2008.239}.

\bibitem[Bennin et~al.(2018)Bennin, Keung, Phannachitta, Monden, and Mensah]{9}
K.~E. Bennin, J.~Keung, P.~Phannachitta, A.~Monden, and S.~Mensah.
\newblock {MAHAKIL}: Diversity based oversampling approach to alleviate the class imbalance issue in software defect prediction.
\newblock In \emph{Proceedings of the 40th International Conference on Software Engineering}, pages 699--699, 2018.
\newblock \doi{10.1145/3180155.3182520}.

\bibitem[He et~al.(2008)He, Bai, Garcia, and Li]{10}
Haibo He, Yang Bai, Edwardo~A Garcia, and Shutao Li.
\newblock {ADASYN}: Adaptive synthetic sampling approach for imbalanced learning.
\newblock In \emph{2008 IEEE International Joint Conference on Neural Networks (IEEE World Congress on Computational Intelligence)}, pages 1322--1328. IEEE, 2008.

\bibitem[Thakur et~al.(2024)Thakur, Jadeja, and Chouhan]{11}
P.~S. Thakur, M.~Jadeja, and S.~S. Chouhan.
\newblock {CBReT}: A cluster-based resampling technique for dealing with imbalanced data in code smell prediction.
\newblock \emph{Knowledge-Based Systems}, 286:\penalty0 111390, 2024.
\newblock \doi{10.1016/j.knosys.2024.111390}.

\bibitem[Xie et~al.(2021)Xie, Liu, Zeng, Lin, and Li]{12}
X.~Xie, H.~Liu, S.~Zeng, L.~Lin, and W.~Li.
\newblock A novel progressively undersampling method based on the density peaks sequence for imbalanced data.
\newblock \emph{Knowledge-Based Systems}, 213:\penalty0 106689, 2021.
\newblock \doi{10.1016/j.knosys.2020.106689}.

\bibitem[Tao et~al.(2025)Tao, Xu, Shi, Li, Guo, and Tao]{13}
X.~Tao, A.~Xu, L.~Shi, J.~Li, X.~Guo, and S.~Tao.
\newblock Density clustering hypersphere-based self-adaptively oversampling algorithm for imbalanced datasets.
\newblock \emph{Knowledge-Based Systems}, 329:\penalty0 114407, 2025.
\newblock \doi{10.1016/j.knosys.2025.114407}.

\bibitem[Ren et~al.(2022)Ren, Zhu, Kang, Fu, Niu, Gao, Yan, and Hong]{14}
Z.~Ren, Y.~Zhu, W.~Kang, H.~Fu, Q.~Niu, D.~Gao, K.~Yan, and J.~Hong.
\newblock Adaptive cost-sensitive learning: Improving the convergence of intelligent diagnosis models under imbalanced data.
\newblock \emph{Knowledge-Based Systems}, 241:\penalty0 108296, 2022.
\newblock \doi{10.1016/j.knosys.2022.108296}.

\bibitem[Khan et~al.(2018)Khan, Hayat, Bennamoun, Sohel, and Togneri]{15}
Salman~H Khan, Munawar Hayat, Mohammed Bennamoun, Ferdous~A Sohel, and Roberto Togneri.
\newblock Cost-sensitive learning of deep feature representations from imbalanced data.
\newblock \emph{IEEE Transactions on Neural Networks and Learning Systems}, 29\penalty0 (8):\penalty0 3573--3587, 2018.
\newblock \doi{10.1109/TNNLS.2017.2732482}.

\bibitem[Rodr{\'\i}guez et~al.(2020)Rodr{\'\i}guez, D{\'\i}ez-Pastor, Arnaiz-Gonz{\'a}lez, and Kuncheva]{16}
Juan~J Rodr{\'\i}guez, Jos{\'e}-Francisco D{\'\i}ez-Pastor, {\'A}lvar Arnaiz-Gonz{\'a}lez, and Ludmila~I Kuncheva.
\newblock Random balance ensembles for multiclass imbalance learning.
\newblock \emph{Knowledge-Based Systems}, 193:\penalty0 105434, 2020.
\newblock \doi{10.1016/j.knosys.2019.105434}.

\bibitem[Zhu et~al.(2023)Zhu, Liu, and Zhu]{17}
T.~Zhu, X.~Liu, and E.~Zhu.
\newblock Oversampling with reliably expanding minority class regions for imbalanced data learning.
\newblock \emph{IEEE Transactions on Knowledge and Data Engineering}, 35\penalty0 (6):\penalty0 6167--6181, 2023.
\newblock \doi{10.1109/TKDE.2022.3171706}.

\bibitem[Li et~al.(2024{\natexlab{a}})Li, Song, Wu, Hu, Cheung, and Yao]{18}
S.~Li, L.~Song, X.~Wu, Z.~Hu, Y.~Cheung, and X.~Yao.
\newblock Multi-class imbalance classification based on data distribution and adaptive weights.
\newblock \emph{IEEE Transactions on Knowledge and Data Engineering}, 36\penalty0 (10):\penalty0 5265--5279, 2024{\natexlab{a}}.
\newblock \doi{10.1109/TKDE.2024.3384961}.

\bibitem[Han et~al.(2005)Han, Wang, and Mao]{19}
Hui Han, Wen-Yuan Wang, and Bing-Huan Mao.
\newblock {Borderline-SMOTE}: A new over-sampling method in imbalanced data sets learning.
\newblock In \emph{Advances in Intelligent Computing}, pages 878--887. Springer, 2005.
\newblock \doi{10.1007/11538059_91}.

\bibitem[Barua et~al.(2014)Barua, Islam, Yao, and Murase]{20}
Sukarna Barua, Md~Monirul Islam, Xin Yao, and Kazuyuki Murase.
\newblock {MWMOTE—Majority} weighted minority oversampling technique for imbalanced data set learning.
\newblock \emph{IEEE Transactions on Knowledge and Data Engineering}, 26\penalty0 (2):\penalty0 405--425, 2014.
\newblock \doi{10.1109/TKDE.2012.232}.

\bibitem[Sun et~al.(2025)Sun, Li, and Zhu]{10902063}
Hao Sun, Jianping Li, and Xiaoqian Zhu.
\newblock A novel expandable borderline smote over-sampling method for class imbalance problem.
\newblock \emph{IEEE Transactions on Knowledge and Data Engineering}, 37\penalty0 (5):\penalty0 2183--2199, 2025.
\newblock \doi{10.1109/TKDE.2025.3544284}.

\bibitem[Li et~al.(2024{\natexlab{b}})Li, Zhou, Liu, Gong, and Wang]{LI2024122593}
Min Li, Hao Zhou, Qun Liu, Xu~Gong, and Guoyin Wang.
\newblock Wrnd: A weighted oversampling framework with relative neighborhood density for imbalanced noisy classification.
\newblock \emph{Expert Systems with Applications}, 241:\penalty0 122593, 2024{\natexlab{b}}.
\newblock \doi{10.1016/j.eswa.2023.122593}.

\bibitem[Liu et~al.(2023)Liu, Liu, Yu, Zhong, and Hu]{21}
Y.~Liu, Y.~Liu, B.~X.~B. Yu, S.~Zhong, and Z.~Hu.
\newblock Noise-robust oversampling for imbalanced data classification.
\newblock \emph{Pattern Recognition}, 133:\penalty0 109008, 2023.
\newblock \doi{10.1016/j.patcog.2022.109008}.

\bibitem[Shen et~al.(2024)Shen, Li, Huan, Shang, Wang, and Chen]{22}
S.~Shen, Z.~Li, Z.~Huan, F.~Shang, Y.~Wang, and Y.~Chen.
\newblock Neighborhood repartition-based oversampling algorithm for multiclass imbalanced data with label noise.
\newblock \emph{Neurocomputing}, 600:\penalty0 128090, 2024.
\newblock \doi{10.1016/j.neucom.2024.128090}.

\bibitem[Xia et~al.(2023)Xia, Shao, Xia, Xiong, Lian, and Ling]{23}
T.~Xia, Y.~Shao, S.~Xia, Y.~Xiong, X.~Lian, and W.~Ling.
\newblock {GBSMOTE}: A robust sampling method based on granular-ball computing and smote for class imbalance.
\newblock In \emph{Proceedings of the 2023 8th International Conference on Mathematics and Artificial Intelligence}, pages 19--24. ACM, 2023.
\newblock \doi{10.1145/3594300.3594304}.

\bibitem[Wang and Yao(2012)]{6170916}
Shuo Wang and Xin Yao.
\newblock Multiclass imbalance problems: Analysis and potential solutions.
\newblock \emph{IEEE Transactions on Systems, Man, and Cybernetics, Part B (Cybernetics)}, 42\penalty0 (4):\penalty0 1119--1130, 2012.
\newblock \doi{10.1109/TSMCB.2012.2187280}.

\bibitem[Tang et~al.(2024)Tang, Li, Hou, Fu, and Tian]{25}
J.~Tang, Y.~Li, Z.~Hou, S.~Fu, and Y.~Tian.
\newblock Robust two-stage instance-level cost-sensitive learning method for class imbalance problem.
\newblock \emph{Knowledge-Based Systems}, 300:\penalty0 112143, 2024.
\newblock \doi{10.1016/j.knosys.2024.112143}.

\bibitem[Yousefimehr et~al.(2026)Yousefimehr, Ghatee, Fazli, Ghaffari, Rafei, Seifi, Tavakoli, Nikahd, Razi~Gandomani, Orouji, Mahmoudi~Kashani, Heshmati, and Mousavi]{make8070211}
Behnam Yousefimehr, Mehdi Ghatee, Javad Fazli, Shervin Ghaffari, Zahra Rafei, Mohammad~Amin Seifi, Sajed Tavakoli, Abolfazl Nikahd, Mahdi Razi~Gandomani, Alireza Orouji, Ramtin Mahmoudi~Kashani, Sarina Heshmati, and Negin~Sadat Mousavi.
\newblock Data balancing strategies: A systematic survey of resampling and augmentation methods.
\newblock \emph{Machine Learning and Knowledge Extraction}, 8\penalty0 (7):\penalty0 211, 2026.
\newblock \doi{10.3390/make8070211}.

\bibitem[Seiffert et~al.(2010)Seiffert, Khoshgoftaar, Van~Hulse, and Napolitano]{26}
Chris Seiffert, Taghi~M Khoshgoftaar, Jason Van~Hulse, and Amri Napolitano.
\newblock {RUSBoost}: A hybrid approach to alleviating class imbalance.
\newblock \emph{IEEE Transactions on Systems, Man, and Cybernetics-Part A: Systems and Humans}, 40\penalty0 (1):\penalty0 185--197, 2010.
\newblock \doi{10.1109/TSMCA.2009.2029559}.

\bibitem[Li et~al.(2026)Li, Chen, and Li]{LI2026129546}
Chuantao Li, Baoqin Chen, and Sheng Li.
\newblock Graph discriminative dynamic sampling adaboost for imbalanced cardiovascular disease diagnosis.
\newblock \emph{Expert Systems with Applications}, 297:\penalty0 129546, 2026.
\newblock \doi{10.1016/j.eswa.2025.129546}.

\bibitem[Datta et~al.(2020)Datta, Nag, and Das]{8621023}
Shounak Datta, Sayak Nag, and Swagatam Das.
\newblock Boosting with lexicographic programming: Addressing class imbalance without cost tuning.
\newblock \emph{IEEE Transactions on Knowledge and Data Engineering}, 32\penalty0 (5):\penalty0 883--897, 2020.
\newblock \doi{10.1109/TKDE.2019.2894148}.

\bibitem[Jan et~al.(2022)Jan, Munos, and Ali]{27}
Z.~Jan, J.~C. Munos, and A.~Ali.
\newblock A novel method for creating an optimized ensemble classifier by introducing cluster size reduction and diversity.
\newblock \emph{IEEE Transactions on Knowledge and Data Engineering}, 34\penalty0 (7):\penalty0 3072--3081, 2022.
\newblock \doi{10.1109/TKDE.2020.3025173}.

\bibitem[Freund and Schapire(1997)]{28}
Yoav Freund and Robert~E Schapire.
\newblock A decision-theoretic generalization of on-line learning and an application to boosting.
\newblock \emph{Journal of Computer and System Sciences}, 55\penalty0 (1):\penalty0 119--139, 1997.
\newblock \doi{10.1006/jcss.1997.1504}.

\bibitem[Zazzaro(2025)]{29}
Gaetano Zazzaro.
\newblock Cross-testing methodology for pattern learning and model transfer in rare fog events detection.
\newblock \emph{Pattern Recognition}, 165:\penalty0 111649, 2025.
\newblock ISSN 0031-3203.
\newblock \doi{10.1016/j.patcog.2025.111649}.

\bibitem[Alcal{\'a}-Fdez et~al.(2011)Alcal{\'a}-Fdez, Fern{\'a}ndez, Luengo, Derrac, Garc{\'i}a, S{\'a}nchez, and Herrera]{32}
Jes{\'u}s Alcal{\'a}-Fdez, Alberto Fern{\'a}ndez, Juli{\'a}n Luengo, Joaqu{\'i}n Derrac, Salvador Garc{\'i}a, Luciano S{\'a}nchez, and Francisco Herrera.
\newblock Keel data-mining software tool: Data set repository, integration of algorithms and experimental analysis framework.
\newblock \emph{Journal of Multiple-Valued Logic and Soft Computing}, 17\penalty0 (2-3):\penalty0 255--287, 2011.

\end{thebibliography}
\end{document}